\documentclass[twocolumn]{IEEEtran}

\usepackage[pdftex]{graphicx} % 插图命令 \includegraphics
\usepackage{amsmath} % Math Packages
\usepackage{algorithmic}
\usepackage{array} % Alignment Packages
\usepackage[caption=false,font=footnotesize]{subfig} % Subfigure Packages
\usepackage{dblfloatfix}
\usepackage{url} % simple URL typesetting

\usepackage{microtype}
\usepackage[utf8]{inputenc} % allow utf-8 input
\usepackage[T1]{fontenc}    % use 8-bit T1 fonts
\usepackage{paralist} % Compacted versions of enumerate and itemize.
\usepackage{marvosym} % 常用符号, 特别是打勾、打叉
\usepackage{bm} % 实现字体的斜体加粗
\usepackage{bbm} % fix \mathbb doesn't support digits
\usepackage{amssymb}   % 黑板粗体 \mathbb, 德文尖角体 \mathfrak, \backsim
\usepackage{mathrsfs} % 花体 \mathscr
\usepackage{tabularray} % tblr or talltblr or longtblr environment
\UseTblrLibrary{booktabs}       % professional-quality tables，也就是三线表
\usepackage{tabularx} % 自动计算列宽的表格包
\usepackage[flushleft]{threeparttable} % 给表格项添加注释
\usepackage{colortbl} % 表格行列颜色 rowcolor
\usepackage{makecell} % 表格内容换行
\usepackage{multirow} % 多行表格
\usepackage{nicefrac} % 斜着的分号
\usepackage[dvipsnames]{xcolor} % 字体颜色
\usepackage{soul} % highlight, strikeout, underline
\setul{0.5ex}{0.3ex}

\definecolor{citecolor}{RGB}{34,139,34}
\usepackage[pagebackref=false,breaklinks=true,letterpaper=true,colorlinks,
    citecolor=citecolor,bookmarks=false]{hyperref}
\usepackage{cleveref}  % for \cref
\crefformat{figure}{Fig.~#2#1#3}
\crefformat{equation}{Eq.~(#2#1#3)}
\crefformat{table}{Table~#2#1#3}
\crefformat{section}{Section~#2#1#3}
\crefformat{algorithm}{Algorithm~#2#1#3}
\crefformat{appendix}{Appendix #2#1#3}
\crefformat{subsection}{subsection~#2#1#3}
\usepackage{cellspace}
\usepackage[numbers,sort&compress]{natbib} % 代替 cite.sty

\usepackage{xspace} % 用于在 \newcommand 中添加空格

\usepackage{pifont} % http://ctan.org/pkg/pifont
\newcommand{\xmark}{\ding{53}}%
\definecolor{Gray}{rgb}{0.9,0.9,0.9}
\definecolor{LightCyan}{rgb}{0.88,1,1}
\newcolumntype{a}{>{\columncolor{Gray}}c}
\newcolumntype{b}{>{\columncolor{white}}c}

\begin{document}

\setlength{\abovedisplayskip}{.5\baselineskip} % 调整公式与正文间的段前距离
\setlength{\belowdisplayskip}{.5\baselineskip} % 调整公式与正文间的段后距离

% \title{Beyond Hit-Miss Trade-Offs: Reshaping Attention with Selective Rank-Aware Squeeze for Infrared \\  Small Target Detection}
% \title{Many Are Called, but Few Are Chosen: Detecting Infrared Small Targets Beyond Hit-Miss Trade-Offs via Selective Rank-Aware Attention}
\title{Pick of the Bunch: Detecting Infrared Small Targets Beyond Hit-Miss Trade-Offs via Selective Rank-Aware Attention}
% \title{Pick of the Bunch: Infrared Small Target Detection Beyond Hit-Miss Trade-Offs via Selective Rank-Aware Attention}

% \title{Beyond Hit-Miss Trade-Offs: Reshaping Attention with Differential Rank-Aware Squeeze for \\ Infrared Small Target Detection}
% \title{Beyond Hit-Miss Trade-Offs: Reshaping Attention via Sparse Squeeze with Constant Complexity for Infrared Small Target Detection}

% !TEX root = ../main.tex

\author{
  Yimian~Dai,
  Peiwen~Pan,
  Yulei~Qian,  
  Yuxuan~Li,
  Xiang~Li,
  Jian~Yang,
  Huan~Wang
  \thanks{
    This work was supported by
      % 青基
      the National Natural Science Foundation of China (62301261, % 我
      62206134, % 翔哥
      61703209, % 王欢
      62272231, % 王欢
      62361166670), % 杨健老师
    % 博后面上
    the Fellowship of China Postdoctoral Science Foundation (2021M701727). % 我
    \emph{(The first two authors contributed equally to this work.
      Corresponding author: Jian Yang and Huan Wang.)}
    }

  % 南理工
  \thanks{
    Yimian Dai and Jian Yang are with PCA Lab, Key Lab of Intelligent Perception and Systems for High-Dimensional Information of Ministry of Education, and Jiangsu Key Lab of Image and Video Understanding for Social Security, while   Peiwen Pan and Huan Wang are with the department of intelligence science, School of Computer Science and Engineering, Nanjing University of Science and Technology, Nanjing, China.
    (e-mail:
    \href{mailto:yimian.dai@gmail.com}{yimian.dai@gmail.com};
    \href{mailto:121106022690@njust.edu.cn}{121106022690@njust.edu.cn};
    \href{mailto:csjyang@mail.njust.edu.cn}{csjyang@mail.njust.edu.cn};
    \href{mailto:wanghuanphd@njust.edu.cn}{wanghuanphd@njust.edu.cn}).
  }

  % 724
  \thanks{Yulei Qian is with Nanjing Marine Radar Institute, Nanjing, China.
  (e-mail: \href{mailto:yuleifly@126.com}{yuleifly@126.com}).
  }
  
  % 南开
  \thanks{Yuxuan Li and Xiang Li are with IMPlus@PCALab \& VCIP, CS, Nankai University.
  Xiang Li also holds a position at the NKIARI, Shenzhen Futian.
  (e-mail:
  \href{mailto:yuxuan.li.17@ucl.ac.uk}{yuxuan.li.17@ucl.ac.uk};
  \href{mailto:xiang.li.implus@nankai.edu.cn}{xiang.li.implus@nankai.edu.cn}).
  }
}

\maketitle

% !TEX root = ../main.tex

\begin{abstract}
  Infrared small target detection faces the inherent challenge of precisely localizing dim targets amidst complex background clutter.
  Traditional approaches struggle to balance detection precision and false alarm rates.
  To break this dilemma, we propose SeRankDet, a deep network that achieves high accuracy beyond the conventional hit-miss trade-off, by following the ``Pick of the Bunch'' principle.
  At its core lies our Selective Rank-Aware Attention (SeRank) module, employing a non-linear Top-K selection process that preserves the most salient responses, preventing target signal dilution while maintaining constant complexity.
  Furthermore, we replace the static concatenation typical in U-Net structures with our Large Selective Feature Fusion (LSFF) module, a dynamic fusion strategy that empowers SeRankDet with adaptive feature integration, enhancing its ability to discriminate true targets from false alarms. 
  The network's discernment is further refined by our Dilated Difference Convolution (DDC) module, which merges differential convolution aimed at amplifying subtle target characteristics with dilated convolution to expand the receptive field, thereby substantially improving target-background separation.
  Despite its lightweight architecture, the proposed SeRankDet sets new benchmarks in state-of-the-art performance across multiple public datasets.
  The code is available at \url{https://github.com/GrokCV/SeRankDet}.
\end{abstract}

% \vspace{-1\baselineskip}
% !TEX root = ../main.tex
% \bibliography{../reference.bib}

\section{Introduction} \label{sec:intro}

% 1. 研究背景的意义
\IEEEPARstart{I}{nfrared} imaging, an unobtrusive thermal sensing technology that captures emitted radiation, offers unparalleled capabilities for discerning distant targets through visual impediments like haze and smoke, independent of light conditions \cite{TGRS2016TIRReview}.
This technology is crucial across military and civilian domains, including surveillance, border security, and search and rescue missions.
Therefore, accurately detecting remote small infrared targets is critical for these applications \cite{PR2023IRSTDSurvey}.

% 2. 所面临的挑战/关键技术瓶颈/关键科学问题
While infrared imaging stands as a powerful tool in discerning remote targets, the nuanced detection of infrared small targets is thwarted by two prevalent challenges that significantly curtail its efficacy in practical applications \cite{GRSM2022SingleFrameSurvey}:
\begin{itemize}
    \item[1)] \textbf{Lack of Distinctive Features:} Small targets often lack sufficient distinctive features, making it difficult to differentiate them from visually similar false alarms using only local target information.
    \item[2)] \textbf{Low Signal-to-Clutter Ratio:} The typically low contrast of infrared small targets against cluttered backgrounds complicates their discernibility, as their signal-to-clutter ratio is minimal, presenting a significant detection hurdle.
\end{itemize}
Therefore, addressing these challenges necessitates enhancing intrinsic target features and integrating contextual analysis to effectively suppress background clutter.

\begin{figure*}[hbtp]
\vspace{-1\baselineskip}
  \centering
  \includegraphics[width=0.98\linewidth]{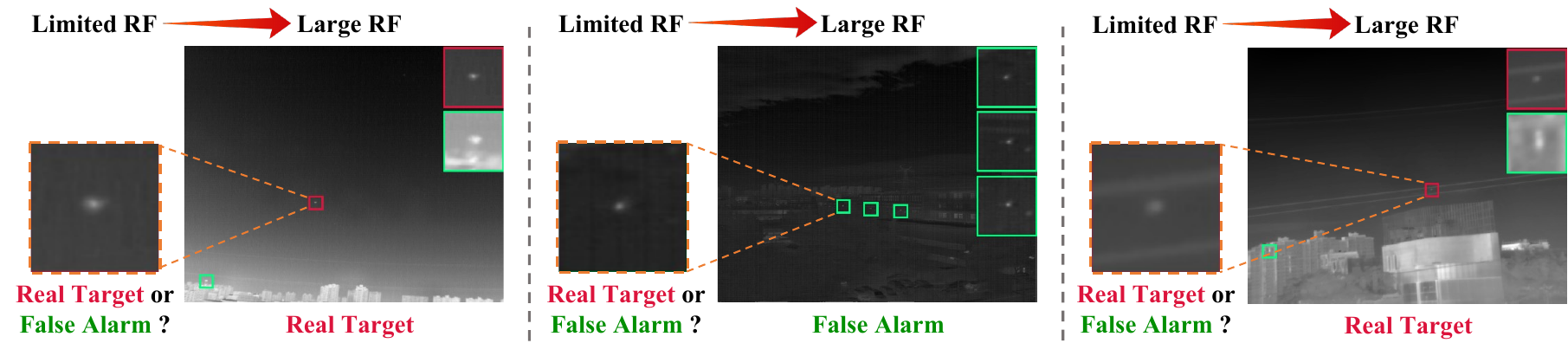}
  % \vspace{-.5\baselineskip}
  \caption{The indispensable role of large background context in infrared small target detection. Merely relying on the cropped region surrounding the target is insufficient to distinguish genuine targets from visually similar false alarms, highlighting the necessity of incorporating extensive contextual information.}
  \label{fig:gallery}
\vspace{-1\baselineskip}
\end{figure*}

% 3. 国内外研究现状
Tackling this problem demands a holistic perspective that integrates multi-scale visual analysis for real-false target differentiation \cite{TGRS2021ALCNet}.
To this end, the research landscape is punctuated by a plethora of deep learning strategies that prioritize the fusion of multi-level contextual information \cite{TGRS2023OSCAR}, aiming to bridge the gap between detailed target characteristics and the broader scene context.
Recent developments include asymmetric contextual modulation techniques that enrich semantic understanding and preserve detail at deeper network layers \cite{dai2021asymmetric}.
Moreover, other strategies employ densely interconnected structures to maintain the integrity of targets, facilitating robust feature interactions across scales \cite{TIP2023DNANet}.

% 4. 之前方法尚未解决的问题
Despite significant strides in infrared small target detection, an unresolved tension at the foundation of feature extraction hampers the attainment of high accuracy. In our opinion, current methods operate more as tactical remedies rather than addressing the central challenges at their root. The shortcomings of existing approaches are manifest in several key areas:
\begin{enumerate}
    \item \textbf{Insensitive Convolutional Layers:} 
    The primary convolutional layers, despite their ubiquity in network frameworks, lack the sensitivity required for the fine-grained details of infrared small targets \cite{TPAMI2021CDC}, leading to insufficient distinction between the targets and complex backgrounds, which is essential for precise detection.
    \item \textbf{Linear Squeeze Computation:} 
    Traditional attention mechanisms, utilizing linear computations akin to the pooling in Squeeze-and-Excitation Networks (SENet) \cite{TPAMI2020SENet} and Pyramid Vision Transformer (PVT) \cite{ICCV2021PVT}, inadvertently merge target features with dominant background noise, diluting the target's signature within the background.
    \item \textbf{Static Feature Fusion:} 
    The prevalent use of static concatenation in network designs, limited to fixed-weight fusion, does not dynamically integrate salient features \cite{WACV2021AFF}. Such rigidity falls short in differentiating true targets from false positives, which is crucial for reliable detection.
\end{enumerate}

Confronted with the dilemma of balancing hit and miss trade-offs \cite{ICCV19Infrared}, we are inspired to ponder: is there a paradigm that can harmoniously reconcile these seemingly contradictory objectives?
This predicament invites a critical inquiry: \textit{can we devise a method that, akin to a discerning connoisseur, meticulously ``\textbf{picks the bunch}'' -- leveraging a highly sensitive feature extractor to preserve even the dimmest of targets, while employing a reliable module to judiciously filter out false alarms?}

% 5. 本文创新 1

Our answer is affirmative. 
To navigate the intricate balance of enhancing detection while mitigating false positives, we introduce SeRankDet, consisting of a trident of modules, which embodies a transformative shift from the conventional paradigm \cite{PR2023IRSTDSurvey}, underscoring the cardinal importance of atomically designing network blocks specifically tailored to this task.

Firstly, to amplify sensitivity towards dim targets, we introduce the \textbf{Dilated Difference Convolution (DDC) }module, a cornerstone in redefining the architecture of the host network.
The DDC module integrates a trio of processes: standard convolution to maintain baseline detection capabilities, differential convolution to accentuate low-intensity target features, and dilated convolution to broaden the receptive field.
This multi-pronged approach empowers the DDC to reliably identify dim targets against background noise with similar characteristics, thereby substantially curtailing false positives.

% 6. 本文创新 2
Secondly, as illustrated in Fig.~\ref{fig:gallery}, the successful discrimination of genuine targets from false alarms necessitates a holistic understanding of the extensive background context, rather than relying solely on localized image patches surrounding the targets.
Inspired by this insight, we develop the \textbf{Large Selective Feature Fusion (LSFF)} module to supplant the static concatenation within the U-Net architecture, constituting the host network for SeRankDet. The LSFF module incorporates a dynamic fusion strategy that transcends traditional static feature concatenation, opting for an expansive selective receptive field within its attentional feature fusion mechanism. By leveraging this selective, wide-ranging receptive field, the LSFF significantly enhances the network's capacity to differentiate true targets from false positives.

% 7. 本文创新 3

Finally, at the heart of SeRankDet lies the proposed \textbf{Selective Rank-Aware Attention (SeRank)} module.
This component diverges from the conventional linear squeezing found in prevalent attention mechanisms, which often inadvertently blend target features with background noise, thereby diluting the distinctiveness required for precise detection. Instead, the SeRank module utilizes a non-linear Top-K operation that strategically amplifies critical target features, maintaining their salience against complex backgrounds while ensuring the computational complexity remains constant.

% 8. 本文贡献
In summary, we advance the state-of-the-art in infrared small target detection with our contributions as follows:
\begin{enumerate}
  \item 
  \textbf{SeRankDet}, a novel network meticulously tailored to the ``Pick of the Bunch'' concept, is introduced. Its core components are atomically re-engineered to effectively detect infrared small targets, circumventing traditional trade-offs between detection accuracy and false positives.
  \item The \textbf{SeRank} module, featuring a Top-K operation, focuses on rank-aware feature aggregation, facilitating an attention mechanism with fixed computational complexity. 
  To our knowledge, this is the first to explore such Top-K squeeze in attention mechanisms.
  \item Our \textbf{DDC} module synergistically combines differential and dilated convolutions with vanilla convolution. This architecture enhances the capture of fine details and simultaneously suppresses background noise through its broad receptive field.
  \item The \textbf{LSFF} module, with its dynamic fusion approach, significantly refines the host network's ability to discriminate against false positives.

\end{enumerate}

% !TEX root = ../main.tex
% \bibliography{../reference.bib}

\section{Related Work} \label{sec:related}

\subsection{Infrared Small Target Detection}

In the field of infrared small target detection, methods diverge into classic signal processing techniques and emerging deep learning paradigms \cite{PR2023IRSTDSurvey}.
Classical techniques typically treat single-frame detection as an anomaly within a quasi-static background, applying local contrast metrics or sparse-low-rank decomposition for target-background separation \cite{TIP20TNLRS}.
Local contrast methods, albeit varied, generally rely on pre-set cellular structures and simple intensity or entropy metrics \cite{GRSL2019LocalContrastTDLMS}, which can lead to imprecise target portrayal.
Conversely, sparse plus low-rank schemes exploit background self-similarity to isolate targets but often falter in differentiating real targets from similar intrusions \cite{TGRS20TopHatReg}, owing to their reliance on low-level features.

With the advent of publicly accessible datasets \cite{TGRS2021ALCNet}, deep learning has surged to the forefront of research in this domain.
Recent advancements showcase varied strategies: Wang \emph{et al.}\ fused reinforcement learning with pyramid features and boundary attention to mitigate localized noise effects \cite{TGRS2023RLPGB}.
Tong \emph{et al.}\ combined encoder-decoder structures, spatial pyramid pooling, and dual-attention with multi-scale labeling to capture the intricate details of target edges and interiors \cite{TGRS2023MSAFFNet}.
Wang \emph{et al.}\ leveraged pixel correlations through a region proposal network and transformer encoder, refined by segmentation for greater precision \cite{TGRS2022IAANet}. 
Li \emph{et al.}\ addressed information loss from pooling with their dense nested attention Network, promoting interaction among hierarchical features to preserve target detail \cite{TIP2023DNANet}. 
Zhang \emph{et al.}'s Dim2Clear, anchored by a context mixer decoder, employed a spatial and frequency attention module to refine feature integration and enhance target visibility \cite{TGRS2023Dim2Clear}.

Note that our work extends beyond existing works on attention mechanisms and feature fusion, with distinct motivation:
\begin{enumerate}
    \item \textbf{Preservation through Rank-Aware Squeezing}: 
    Our work distinguishes itself by addressing the common but critical issue of target information loss due to linear squeezing in attention mechanisms. The proposed SeRank module uses a rank-aware Top-K operation to maintain essential target features by effectively preserving their prominence during the squeezing process.
    \item \textbf{Enhanced Early Detection via DDC}: While traditional local contrast methods capture the pop-out characteristics of infrared small targets at the image level, our DDC module operates at the feature level.
    By computing the difference between central and neighboring feature points, the DDC module captures local image brightness changes, enhancing sensitivity to low-contrast, weak small targets in early stages.
\end{enumerate}

\begin{figure*}[hbtp]
    \centering
    \includegraphics[width=0.98\linewidth]{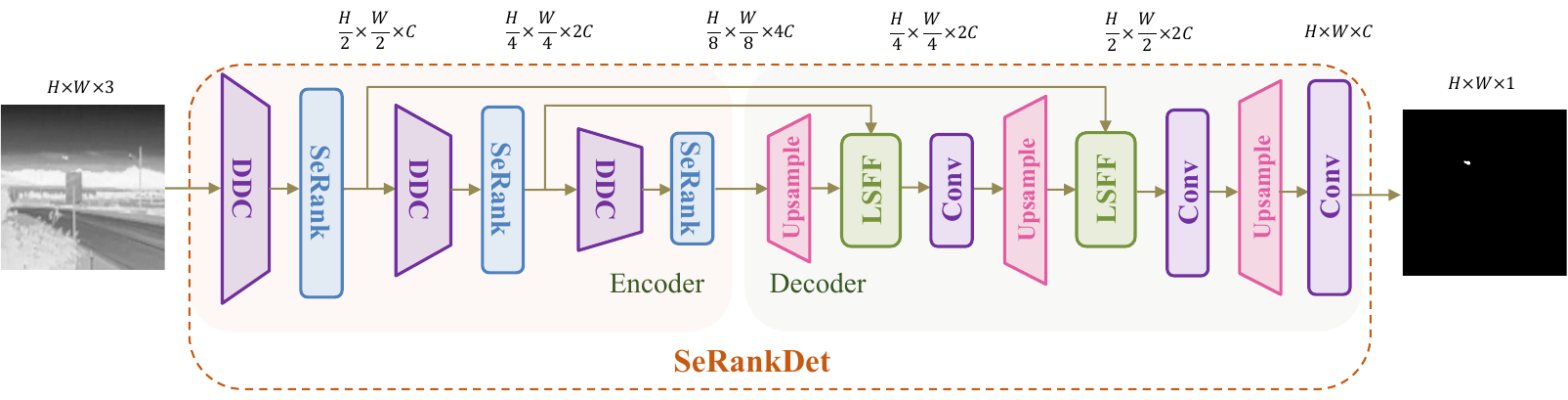}
    \vspace{-.5em}
    \caption{Schematic overview of the proposed SeRankDet network. SeRankDet incorporates three specially designed modules: DDC, SeRank, and LSFF. \textit{For simplicity, only three out of the five stages are depicted.}}
    \label{fig:SeRankDet-Arch}
\end{figure*}

\subsection{Attention Mechanisms}

Attention mechanisms have significantly improved neural network performance in computer vision by dynamically prioritizing salient features \cite{zhang2024global, zhong2024hierarchical}.
The squeeze and excitation (SE) block fine-tuned convolutional neural networks (CNNs) by recalibrating channel-wise features through inter-channel correlation \cite{TPAMI2020SENet}. In addition, spatial attention has been explored to harness long-range dependencies, with the gather-excite network (GENet) employing depth-wise convolutions for feature aggregation \cite{NIPS18GENet}.

Our previous works, the selective kernel network (SKNet) \cite{CVPR2019SKNet} and the Large Selective Kernel Network (LSKNet) \cite{ICCV2023LSKNet,li2024lsknet}, pioneered adaptive receptive field selection.
SKNet introduces a dynamic selection mechanism that allows each neuron to adjust its receptive field size based on multiple scales of input information. LSKNet extends this concept to the domain of remote sensing object detection, leveraging large and selective kernels to model the varying long-range context required by different types of objects.

This work presents the Large Scale Feature Fusion (LSFF) module, an extension of the large receptive field selection concept from LSKNet, but with two key differences:
\begin{enumerate}
    \item \textbf{Same-Layer Fusion vs. Cross-Layer Fusion}: Unlike SKNet and LSKNet, which perform feature fusion within the same layer of the backbone network, the LSFF module integrates large receptive field selection into U-Net's feature fusion via its long-skip connections. This cross-stage fusion harnesses spatial context information from different network depths.
    \item \textbf{Manual Construction of Large Receptive Field vs. Utilization of Intrinsic Large Receptive Field}: While SKNet and LSKNet construct large receptive fields by selectively employing different convolution kernels, the LSFF module capitalizes on U-Net's built-in large receptive fields within its encoder-decoder configuration. This approach eliminates the need for manual sampling of large kernels, thereby simplifying the network architecture and reducing computational overhead, while maintaining robustness to scale variations in the input data.

\end{enumerate}

The success of the transformer architecture has elevated the role of self-attention in vision tasks \cite{zhu2024towards}. 
Non-local Networks initially addressed capturing long-range dependencies with their non-local operations but at a steep computational cost.
Criss-Cross Network (CCNet) optimized this process through criss-cross attention \cite{TPAMI2023CCNet}, which more efficiently accumulates dependencies.
The Pyramid Vision Transformer (PVT) took a different tack for dense prediction tasks, employing a spatial-reduction attention layer that lessens the computational load via pooling operations \cite{ICCV2021PVT}.
Diverging from PVT's approach, our SeRank module is characterized by four distinct features:
\begin{enumerate}

    \item \textbf{Salient Feature Retention}: SeRank is engineered to enhance the preservation of salient features through a non-linear Top-K selection for feature squeezing, providing a nuanced alternative to PVT's linear average pooling.
    \item \textbf{Spatial vs. Channel-wise Self-Attention}: 
    SeRank contrasts PVT by employing channel-wise self-attention, thereby circumventing the need for spatial interpolation and associated resolution scaling.
    \item \textbf{Constant Computational Complexity}: Leveraging Top-K selection and channel-wise self-attention, the SeRank module achieves a computational complexity that is independent of the feature maps' spatial dimensions. This contrasts with PVT, where computational demands escalate with increasing feature map size.
\end{enumerate}

Recently, the integration of Top-K operations into self-attention mechanisms is an emerging approach to enhance efficiency in large language models.
Gupta \emph{et al.}\ optimized memory use with chunk-based query processing that prioritizes key queries, eliminating the need for corrective pre-training \cite{gupta2021memoryefficient}.
Similarly, for vision tasks, QuadTree Attention strategically focuses on pertinent patches, guided by attention scores, for computational frugality \cite{tang2022quadtree}. 
DRSformer introduces a learnable Top-K operator that discerningly aggregates pivotal attention scores, refining feature quality for image deraining applications \cite{CVPR2023DRSformer}.
Our SeRank module, while employing Top-K selection, diverges from these methods in critical aspects:
\begin{enumerate}
    \item \textbf{Distinct Top-K Operator Usage:}
    SeRank uniquely applies Top-K selection directly to the input features as $\mathcal{T}_k(\mathbf{Q})\mathcal{T}_k(\mathbf{K}^\top)$, unlike previous methods that apply it post-attention map computation $\mathcal{T}_k(\mathbf{Q}\mathbf{K}^\top)$, where $\mathbf{Q}$ and $\mathbf{K}$ represent the query and key matrices, and $\mathcal{T}_k(\cdot)$ is the Top-K operator.
   
    \item \textbf{Different Sparse Attention Dimension:} In contrast to the spatial attention used in prior work, SeRank emphasizes channel-wise self-attention, showcasing its difference in handling attention mechanisms.
\end{enumerate}

\subsection{U-Net Variants}

The U-Net architecture has emerged as a cornerstone for image segmentation \cite{MICCAI15UNet}, recognized for its encoder-decoder structure where concatenated feature maps preserve critical detail for precise segmentation. Its exemplary performance has solidified its utilization in various segmentation applications \cite{10376277}, prompting a series of progressive advancements.

Enhancements to U-Net have primarily unfolded along two key dimensions: \emph{Backbone Design Enhancements} and \emph{Skip Connection Enhancements}.
The backbone has evolved with improved encoder blocks; for instance, Residual U-Net leveraged skip connections akin to ResNet \cite{CVPR16ResNetV1}, enabling the construction of deeper architectures \cite{8309343}.
MultiResUNet introduced inception-like modules to the U-Net structure, facilitating multi-scale feature extraction and improving segmentation capabilities \cite{ibtehaz2020multiresunet}. 
DUNet capitalized on deformable convolutions to dynamically adapt receptive fields, while R2U-Net's recurrent residual blocks have shown to markedly improve performance in medical image segmentation \cite{JIN2019149}.
Our DDC module aligns with the objective of improving the network through advanced basic blocks but diverges from existing U-Net variants in several ways:
\begin{enumerate}
    \item \textbf{Tailored for Infrared Small Targets:} 
    The DDC module is meticulously designed for infrared small target feature extraction, finely balancing the trade-off between capturing delicate features and incorporating a wider receptive field for contextual understanding.
    \item \textbf{Unique Blend of Convolutions:}
    Moving beyond the typical enhancements of basic blocks through skip connections, our module amalgamates a trio of convolution types to enhance feature representation.
    To our knowledge, our DDC module is the first of such integration within U-Net in one basic block.
\end{enumerate}

Skip connection enhancements aim to refine feature integration between the encoder and decoder.
U-Net++ leveraged dense connections for versatile multi-scale feature aggregation \cite{UNetPlusPlus}, while U-Net3+ introduced full-scale skip connections to explicitly learn organ positions and boundaries \cite{9053405}.
BiO-Net \cite{xiang2020bio} and BCDU-Net \cite{9022282} both utilized bidirectional connections for feature fusion without increasing parameter count, with BCDU-Net adding convolutional LSTM modules. 
Attention U-Net distinguished itself by incorporating attention gates to emphasize salient features and diminish irrelevant areas \cite{Oktay2018AttentionUL}.

Our LSFF module extends the principles of attention-based feature selection, as seen in Attention U-Net, with two significant differences:
\begin{enumerate}
    \item \textbf{Dynamic Feature Fusion:}
    In Attention U-Net, the attention module merely modulates low-level features, while static concatenation is still used for feature fusion. In contrast, our LSFF module uses attention mechanism as the dynamic feature fusion for skip connection.
    \item \textbf{Pioneering Large Selective Kernel in U-Net:} To the best of our knowledge, this is the first attempt to transplant the idea of large selective kernel from the backbone to the long-skip connection fusion in U-Net.
\end{enumerate}

% !TEX root = ../main.tex
% \bibliography{../reference.bib}

\section{METHODS} \label{sec:method}

\subsection{SeRankDet Architecture}

The overall architecture of the proposed SeRankDet is illustrated in Fig.~\ref{fig:SeRankDet-Arch}, built upon a U-Net-like backbone enhanced by the integration of three key components: the DDC, SeRank, and LSFF modules. 
The encoder pathway consists of five integrated layers, each containing a DDC module followed by a SeRank module. The DDC modules replace vanilla convolutions in U-Net, aiming to accentuate subtle target features and incorporate larger contextual regions. The succeeding SeRank modules perform channel-wise feature squeezing via a nonlinear Top-K operation, spotlighting the most salient target responses to prevent dilution by the background clutter. The squeezed output is then passed to the next integrated layer for further processing in a cascaded manner. The decoder pathway contains four LSFF modules that dynamically fuse encoder features across different scales, transmitted via skip connections. This multi-level feature aggregation helps distinguish true small targets from false alarms more effectively. Finally, the fused decoder features are fed into detection headers to yield the ultimate predictions.

\subsection{Dilated Difference Convolution (DDC) Module}

Conventional convolutional layers in CNNs suffer from a limited receptive field, hindering their ability to capture long-range dependencies within the data. To address this, we introduce the Dilated Difference Convolution (DDC) module, a sophisticated basic block designed to substantially improve the network's feature learning prowess.

\begin{figure}[hbtp]
    \centering
    % \vspace{-1\baselineskip}
    \includegraphics[width=0.98\linewidth]{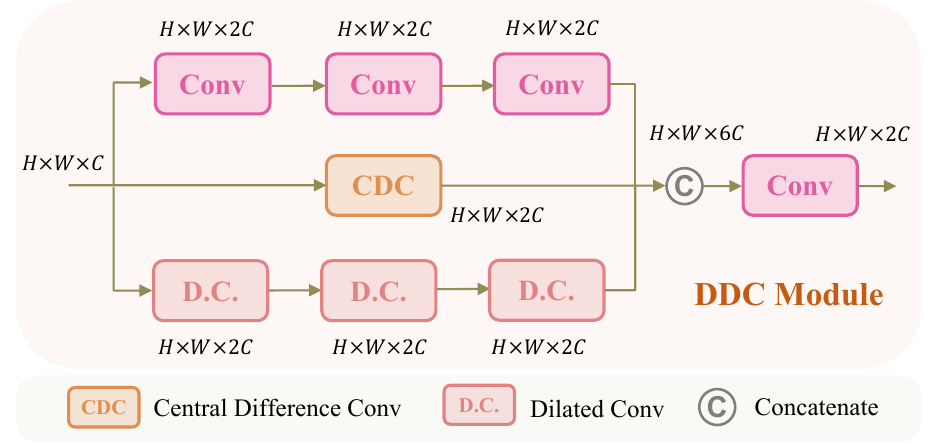}
    \caption{Illustration of the proposed DDC module: Synergizing convolutions to enhance the capture of intricate details while simultaneously suppresses background noise through its broad receptive field.}
    \label{fig:DDC-Module}
\end{figure}

Fig.~\ref{fig:DDC-Module} depicts our DDC module, which integrates three specialized branches operating in parallel for enhanced feature extraction. The first branch employs standard convolution to maintain local spatial coherence, ensuring that essential patterns within close proximity are preserved. The second branch, featuring a differential convolution, is tailored to highlight intricate edge details, critical for distinguishing subtle structures within the image. The third branch utilizes dilated convolution with varying dilation rates of 2, 4, and 2 across its layers, designed to enlarge the receptive field and assimilate contextual information from broader spatial extents. This multi-dilation approach allows the module to grasp context at different scales, enriching the feature set.
The outputs of these branches are then merged, leveraging a $1\times 1$ convolution to synergize the diverse representations into a cohesive feature map.

The DDC module is adept at capturing features across multiple scales, fusing local and long-range information. 
Leveraging a fusion of convolutional operations, it constructs a nuanced feature landscape that is sensitive to both the granular details and broader contextual cues of infrared small targets. This multifaceted approach amplifies the module's ability to discern edges and subtle details, surpassing traditional convolutional methods. With its expanded receptive field, the DDC module demonstrates superior contextual awareness and noise resilience, enhancing the clarity and reliability of feature extraction.

\subsection{Selective Rank-Aware Attention (SeRank) Module}

While the DDC module expands the receptive field to assimilate contextual information, it remains challenged in handling background clutter from distant sources that may be erroneously identified as targets.
Traditional attention mechanisms exacerbate this problem as their linear squeezing operations can dilute critical target features, risking information loss.
To tackle these challenges, we propose the Selective Rank-Aware Attention (SeRank) module, the cornerstone of our SeRankDet framework. SeRank is designed to accentuate target signatures while concurrently attenuating noise, thereby refining the robustness of feature extraction against complex backgrounds.

\begin{figure}[hbtp]
    \centering
    \includegraphics[width=0.98\linewidth]{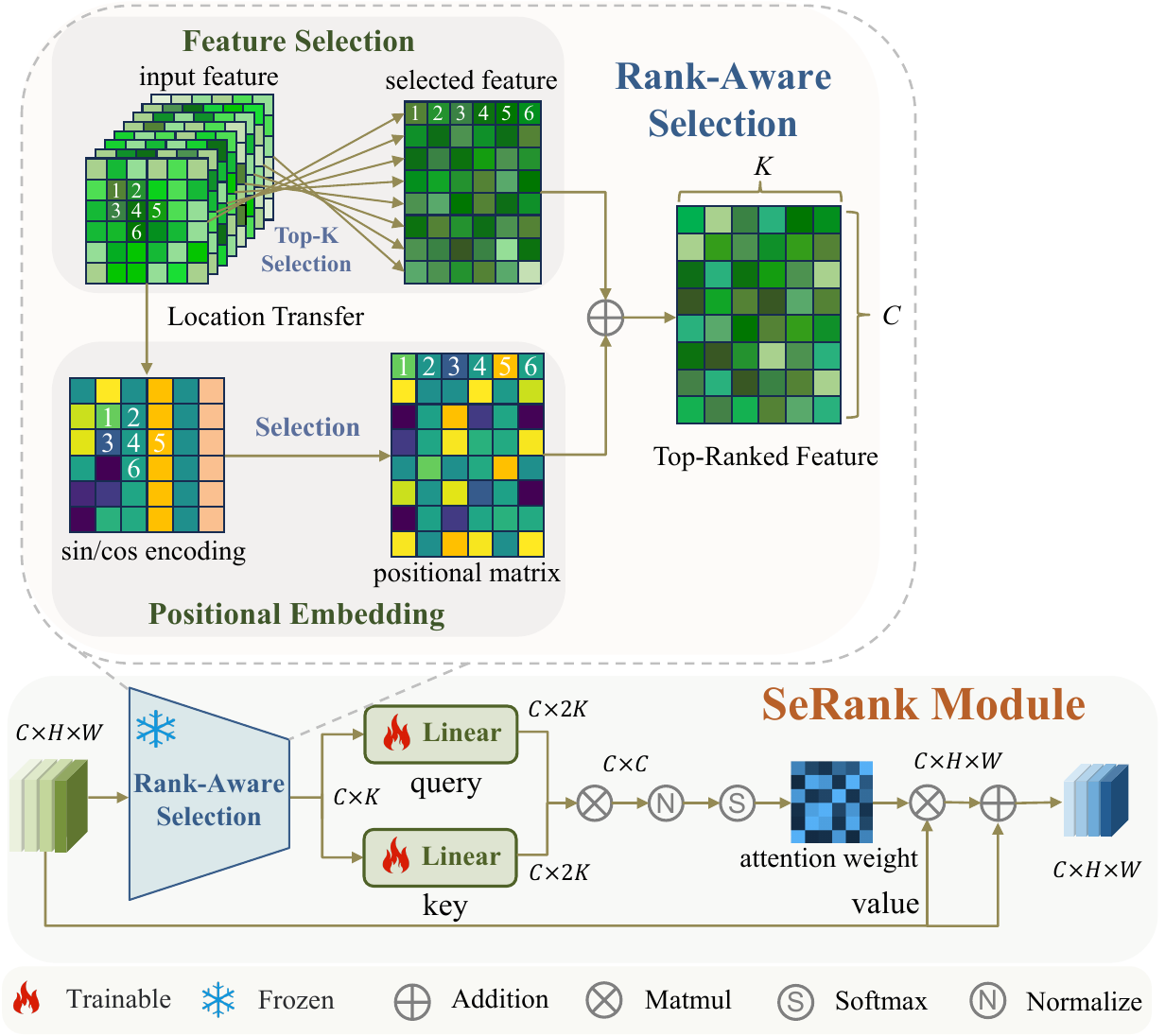}
    \caption{Illustration of the proposed SeRank module, which amplifies target features with Top-K operation. It preserves pivotal target features against complex backgrounds while maintaining constant computational complexity.}
    \label{fig:SeRank-Module}
\end{figure}

Fig.~\ref{fig:SeRank-Module} showcases the SeRank module, which operates through three  stages: \emph{rank-aware feature selection}, \emph{position embedding}, and \emph{channel-wise self-attention}.
Initially, SeRank identifies and isolates the most prominent features across channels, forming a matrix that encapsulates key signatures.
Positional encoding is then integrated to endow these features with positional context \cite{vaswani2017attention}, spotlighting areas of interest for potential targets.
Finally, fully-connected layers generate channel-specific query and key matrices, leading to a channel self-attention mechanism that refines the focus on targets and attenuates noise, resulting in an optimized attention map.

\subsubsection{\textbf{Top-K Feature Selection}}
Our SeRank module employs a Top-K operation to sample the most prominent points in each channel, concatenating them into a feature matrix $\mathbf{F} \in \mathbb{R}^{C\times K}$, where $K$ denotes the number of sampled points dynamically determined based on layer depth.
The Top-K operation is defined as follows:
\begin{equation}
    \renewcommand{\arraystretch}{1.25}
    \begin{array}{cc}
         [f_1,f_2,\dots,f_K], [l_1,l_2,\dots,l_K] =\mathrm{TopK}(\mathbf{X}, k=K),  \\
         \mathbf{F}=\mathrm{Concat}([f_1,f_2,\dots,f_K], \mathrm{dim}=1),\\
    \end{array}
    \label{eq:topk}
    \end{equation}
where $\mathbf{X}\in \mathbb{R}^{C \times H \times W}$ represents the input tensor, $f \in \mathbb{R}^C$ denotes the selected feature vectors, and $l$ represents the corresponding indices.
The resulting feature matrix $\mathbf{F}$ serves as a compact and informative representation of the most salient features extracted from the input tensor, enabling subsequent processing steps to focus on the most discriminative aspects of the data.

The value of $K$ is adaptively determined based on the depth of the network layer, allowing the SeRank module to capture an appropriate number of salient feature points at different scales. We employ the following formulation to compute $K$:
\begin{equation}
K=2^{\lfloor\mathrm{log}_2(C)+o-2(i-1)\rfloor},
\end{equation}
where $C$ is the number of channels in the input tensor, $o$ is a preset offset, and $i$ represents the actual layer number within the network. 
The offset $o$ allows for fine-tuning the number of selected feature points at each layer.
This formulation ensures that deeper layers, which typically have a higher number of channels, sample a larger number of feature points compared to shallower layers. 
By dynamically adjusting $K$ based on layer depth, the SeRank module effectively captures the most informative features at various levels of abstraction throughout the network.

\subsubsection{\textbf{Positional Embedding}}
After re-arranging through feature selection, the original feature points lose their positional information in the feature map. Therefore, we use this module to assign positional information to them. First, we generate a position encoding $\mathbf{E}\in\mathbb{R}^{H\times W}$ with the same width and height as the feature map. We employ the $\sin$ and $\cos$ position encoding for constructing $\mathbf{E}$, with $\sin$ encoding filling the even-numbered columns and $\cos$ encoding filling the odd-numbered columns:
\begin{equation}
    \renewcommand{\arraystretch}{1.5}
    \begin{array}{cc}
         \mathbf{E}[:,0::2]=\sin(pos\times d),  \\
         \mathbf{E}[:,1::2]=\cos(pos\times d), 
    \end{array}
    \label{eq:sincos}
\end{equation}
where $pos=[0,1,\dots,H]$ represents the positions of the input sequence, $d=[0,2,\dots,W]\cdot\mathrm{exp}(-\mathrm{log}(10000)/W)$ is used to compute the sin and cos components in the position encoding.

The $l = [l(x), l(y)]$ in Eq.~(\ref{eq:topk}) represents the two-dimensional coordinate information of the selected feature points in the feature map. Based on $l$, we encode the corresponding positions of each feature point in matrix E to form a position vector $[p_1,p_2,\dots,p_K]$. Then, we concatenate the feature vectors as well and obtain the position matrix $\mathbf{P} \in \mathbb{R}^{C\times K}$:
\begin{equation}
    \label{eq1}
    \renewcommand{\arraystretch}{1.5}
    \begin{array}{cc}
        [p_1,p_2,\dots,p_K] = \mathbf{E}[l(x), l(y)], \\
         \mathbf{P}=\mathrm{Concat}([p_1,p_2,\dots,p_K], \mathrm{dim}=1).
    \end{array}
    \end{equation}
The summed position-aware feature matrix $\mathbf{F}+\mathbf{P}$ encapsulates both salient features and their spatial distribution:
\begin{equation}
    \mathcal{T}_{\mathrm{p}} (\mathbf{X}) = \mathbf{F} + \mathbf{P},
\end{equation}
where $\mathcal{T}{\mathrm{p}}(\mathbf{X}) \in \mathbb{R}^{C\times K}$ is the selected feature matrix.

\begin{figure*}[htbp]
    \centering
    \vspace{-1\baselineskip}
    \includegraphics[width=.98\linewidth]{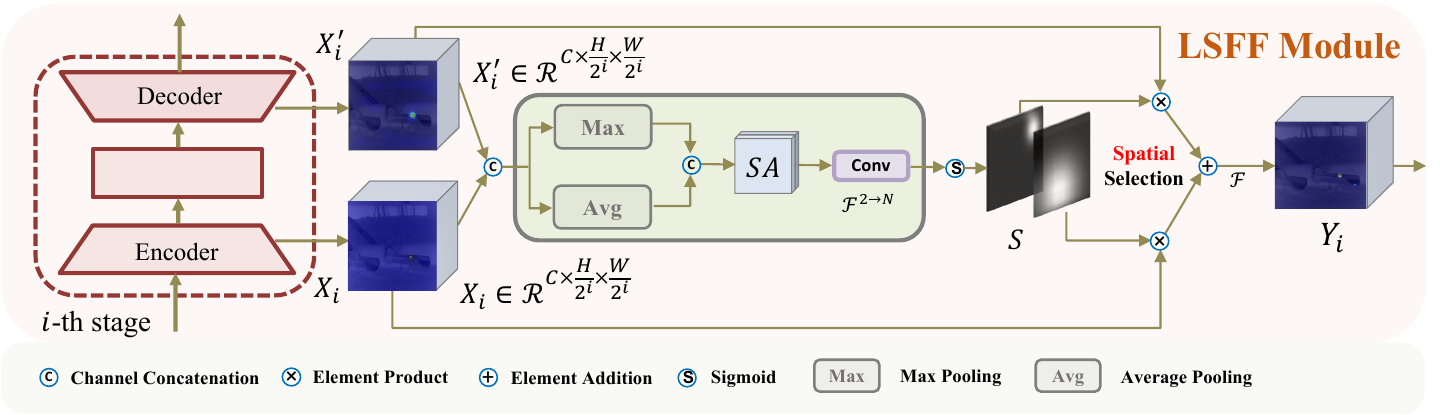}
    \caption{Architecture of the proposed LSFF module: advancing beyond static concatenation in U-Net with dynamic selective feature fusion for better true positive discrimination in the SeRankDet framework.}
    \vspace{-1\baselineskip}
    \label{fig:lsff}
\end{figure*}

\subsubsection{\textbf{Channel-wise Self-Attention}}
As to the attention computation, SeRank first converts $\mathcal{T}{\mathrm{p}}(\mathbf{X})$ into query $\mathbf{Q}_{\mathrm{s}} \in \mathbb{R}^{C\times 2K}$ and key $\mathbf{K}_{\mathrm{s}} \in \mathbb{R}^{C\times 2K}$ matrices via fully connected layers:
\begin{align}
\mathbf{Q}_{\mathrm{s}} & =  \mathcal{T}_{\mathrm{p}}(\mathbf{X}) \bm{W}_Q, \\
\mathbf{K}_{\mathrm{s}} & =  \mathcal{T}_{\mathrm{p}}(\mathbf{X}) \bm{W}_K,
\label{eq:convert}
\end{align}
where the weight matrices $\bm{W}_Q$ and $\bm{W}_K$ have a shape of $K \times 2K$, projecting $\mathcal{T}{\mathrm{p}}(\mathbf{X})$ into a higher-dimensional space.

It is important to emphasize that our choice of implementing channel-wise self-attention in the SeRank module is primarily motivated by the dimensional constraints imposed by the preceding Top-K operation.
The dimensional mismatch prevents the direct application of spatial self-attention. In contrast, along the channel dimension, the dimensions of $\mathbf{Q}_{\mathrm{s}}$, $\mathbf{K}_{\mathrm{s}}$, and $\mathbf{X}$ are consistently $C$. Therefore, following the Top-K operation, we opt to apply self-attention along the channel dimension to circumvent the dimensional incompatibility issue that would arise with spatial self-attention.

We compute the channel self-attention matrix via $\mathbf{Q}_{\mathrm{s}}^{\top} \mathbf{K}_{\mathrm{s}}$, normalize it using softmax to obtain attention scores, and apply the resulting attention map to the input features $\mathbf{X}$, effectively spotlighting target signatures. The output is then fused with the input through a residual connection:
\begin{align}
\operatorname{SeRank}(\mathbf{X}) & =  \left( 1 +  \operatorname{softmax}\left(\frac{\mathbf{Q}_{\mathrm{s}} \mathbf{K}_{\mathrm{s}}^{\top} - \mu}{\sigma + \varepsilon}\right) \right) \mathbf{X},
\label{eq:serank}
\end{align}
where $\mu$ and $\sigma$ are the mean and standard deviation of attention matrix, respectively.

Notably, our SeRank module diverges from conventional self-attention by focusing solely on channel dimensions, thereby achieving a computational complexity of $O(C^2 \cdot K)$, which remains independent of the spatial dimensions $H \times W$. This design not only ensures efficiency but also maintains the module's focus on enhancing feature representation without the overhead of processing spatial dimensions.

\subsection{Large Selective Feature Fusion Module}

Our proposed LSFF module seeks to enhance the network's ability to discriminate between actual infrared small targets and similar false alarms. We achieve this by incorporating a dynamic spatial selection mechanism within the U-Net framework, as depicted in Fig.~\ref{fig:lsff}. This mechanism adeptly exploits multi-scale features, enabling the network to concentrate on the most informative regions of the spatial context.

The LSFF module specifically operates on the feature maps from the encoder-decoder pairs at the i-th stage. The feature maps, denoted as $\mathbf{X}_i, \mathbf{X}^{\prime}_i \in \mathbb{R}^{C_i \times \frac{H}{2^i} \times \frac{W}{2^i}}$, consist of $C_i$ channels and have spatial dimensions of $\frac{H}{2^i}$ and $\frac{W}{2^i}$ in height and width respectively. The feature maps are merged as follows:
\begin{equation}
    \mathbf{U}_i = \mathrm{Concat}\left(\mathbf{X}_i, \mathbf{X}^{\prime}_i\right).
    \label{eqn:4}
\end{equation}
To alleviate the computational burden, we apply channel-wise average and maximum pooling operations (denoted as $\mathcal{P}^{\mathrm{avg}}(\cdot)$ and $\mathcal{P}^{\mathrm{max}}(\cdot)$) to $\mathbf{U}_i$:
\begin{equation}
    \mathbf{SA}^{\mathrm{avg}}_i = \mathcal{P}^{\mathrm{avg}}(\mathbf{U}_i)\text{, \ }
    \mathbf{SA}^{\mathrm{max}}_i = \mathcal{P}^{\mathrm{max}}(\mathbf{U}_i)\text{.}
    \label{eqn:5_2}
\end{equation}
Here, $\mathbf{SA}^{\mathrm{avg}}_i$ and $\mathbf{SA}^{\mathrm{max}}_i$ denote the average and maximum pooled spatial feature descriptors. We concatenate these spatially pooled features to generate an aggregated matrix, thereby enabling information interaction among different spatial descriptors and facilitating the selective fusion process.

Next, we employ a convolution layer $\mathcal{F}^{2 \rightarrow N}(\cdot)$ to transform these aggregated features into $N$ spatial attention maps:
\begin{equation}
    \widehat{\mathbf{SA}}_i = \mathcal{F}^{2 \rightarrow N}(\mathrm{Concat}(\mathbf{SA}^{\mathrm{avg}}_i, \mathbf{SA}^{\mathrm{max}}_i))\text{.}
    \label{eqn:6.2}
    % \vspace{-4pt}
\end{equation}
The final output of the LSFF module is the element-wise product between the input feature $\mathbf{X}_i$ and $ \mathbf{X}^{\prime}_i$:
\begin{equation}
\mathbf{Y}_i = \mathbf{X}_i \odot \sigma(\widehat{\mathbf{SA}}_i) + \mathbf{X}^{\prime}_i \odot  (1 - \sigma(\widehat{\mathbf{SA}}_i)),
\end{equation}
where $\sigma(\cdot)$ signifies the sigmoid function, yielding the dynamic fusion weights. This adaptive method allows the LSFF module to identify and integrate the most salient features, thereby boosting the network's representational capacity.

\begin{table*}[htbp]
  \setlength{\abovecaptionskip}{0cm}  %段前
  \renewcommand\arraystretch{1.2}
  \footnotesize
  \centering
  \vspace{-1\baselineskip}
\caption{CONFIGURATIONS FOR ALL COMPARATIVE EXPERIMENTS}
\vspace{-1\baselineskip}
\label{tab:parameters}
\tabcolsep=0.3cm
\renewcommand\arraystretch{1.2}
\begin{center}
\begin{tabular}{lll}
\toprule[1pt]
Methods               & Source         & Key parameters configuratons   \\ \hline
\multicolumn{3}{l}{\textit{Background Suppression Methods}}  \\ \hline
NWMTH \cite{bai2010analysis} & PR'2010 & $R_{o}=9,R_{i}=4$ \\ 
ILCM \cite{GRSL2014ILCM} & GRSL'2014 & Cell size: $3 \times 3$, threshold $k = 3$ \\
MPCM \cite{PR16MPCM} & PR'2016 & $N = 1, 3, \cdots, 9$, threshold $k = 3$ \\
WLDM \cite{deng2016small} & TGRS'2016 & $L = 4, m = 2, n = 2$, threshold $k = 2$ \\
RLCM \cite{GRSL2018RLCM} & GRSL'2018 & $k_1 = [2,5,9], k_2 = [4, 9, 16]$, scale: 3, threshold $k = 1$ \\
FKRW \cite{qin2019infrared} & TGRS'2019 & Windows size: 11, $K=4,p=6,\beta=200$ \\ 
TLLCM \cite{han2019local} & GRSL'2020 & Scales:[5,7,9] $\lambda =0.5$, threshold $ k=3$ \\ 
GSWLCM \cite{qiu2022global} & GRSL'2022 & Local Window Structures: $[3,5,7,9]$, $\delta=0.01, k = 20$ \\ \hline
\multicolumn{3}{l}{\textit{Low-rank and Sparse Decomposition}}  \\ \hline
IPI \cite{gao2013infrared} & TIP'2013 & Patch size: 50, sliding step: 10, $\lambda=1/\sqrt{\mathrm{min}(m,n)}$, $\epsilon=10^{-7}$, $k = 8$ \\ 
NIPPS \cite{dai2017non} & IPT'2017 & Patch size: 50, sliding step: 10, $\lambda=2/\sqrt{\mathrm{min}(m,n)}$, $\epsilon=10^{-7}$, $k = 3$ \\ 
RIPT \cite{dai2017reweighted} & JSTARS'2017 & Patch size: 50, sliding step: 10, $\lambda=2/\sqrt{\mathrm{min}(m,n)}$, $\epsilon=10^{-2}$, $\omega=10^{-7}$, $k = 5$ \\ 
NRAM \cite{zhang2018infrared}       &  RS'2018     & Patch size: 50, sliding step: 10, $\lambda=1/\sqrt{\mathrm{min}(m,n)}$, $\mu^0=3/\sqrt{\mathrm{min}(m,n)}$, $C=\sqrt{\mathrm{min}(m,n)}/2.5$, $\epsilon=10^{-7}$, $k = 6$ \\ 
NOLC \cite{zhang2019infrared}              &   RS'2019    & Patch size: 30, sliding step: 10, $\lambda=1/\sqrt{\mathrm{max}(size(D))}$, $p=0.5$, $k = 3$ \\ 
PSTNN \cite{zhang2019infrared1}              &    RS'2019   & Patch size: 40, sliding step: 40, $\lambda=0.6/\sqrt{\mathrm{max}(n_1,n_2)\times n_3}$, $\epsilon=10^{-7}$, $k = 2$ \\ \hline
\multicolumn{3}{l}{\textit{Deep Learning Methods}}  \\ \hline
ACM \cite{dai2021asymmetric}             &  WACV'2021     &  Backbone: FPN, layer blocks: [4, 4, 4], channels: [8, 16, 32, 64]                    \\ 
AGPCNet \cite{zhang2021agpcnet}               &   TAES'2023    &  Backbone: resnet34, scales: [10, 6, 5, 3], reduce ratios: [16, 4], gca type: patch, gca att: post                     \\ 
DNANet \cite{TIP2023DNANet}              &  TIP'2022     &    Backbone: resnet18, layer blocks: [2, 2, 2, 2], filter: [16, 32, 64, 128, 256]                   \\
ISNet \cite{zhang2022isnet}               &  CVPR'2022     &  Backbone: resnet18, layer blocks: [4, 4, 4], channels: [8, 16, 32, 64], fuse mode: AsymBi              \\ 
UIUNet \cite{wu2022uiu}              &  TIP'2023     &  Channels: [64, 128, 256, 512], fuse mode: AsymBi                     \\
RDIAN \cite{sun2023receptive}              &   TGRS'2023    &    Channels: [32, 32, 32, 32], layer blocks: [1, 2, 2, 2], kernel size: [1, 3, 5, 7]                   \\

MTUNet \cite{wu2023mtu}               &  TGRS'2023     &  Backbone: resnet18, layer blocks: [2, 2, 2, 2], filter: [16, 32, 64, 128, 256]                     \\ 
 
ABC \cite{pan2023abc}               &  ICME'2023     &    Channels: [64, 128, 256, 512, 1024], dilate rates: [2, 4, 2]                   \\
SeRankDet & Ours & Channels: [64, 128, 256, 512, 1024], $o=3$ \\
\bottomrule[1pt]
\end{tabular}
\end{center}
\end{table*}

% !TEX root = ../main.tex
% \bibliography{../reference.bib}

\section{Experiments} \label{sec:experiment}

\subsection{Experimental Setup} \label{subsec:setup}

\subsubsection{\textbf{Datasets}} \label{subsubsec:dataset}

In the experimental phase of our study, we selected four widely recognized datasets as benchmarks: SIRST \cite{dai2021asymmetric}, IRSTD1K \cite{zhang2022isnet}, SIRSTAUG \cite{zhang2021agpcnet}, and NUDT-SIRST \cite{TIP2023DNANet}. These datasets, renowned for their diversity and complexity, provide a rigorous testing ground for our proposed method, ensuring a comprehensive evaluation of its performance and generalization capabilities.

\begin{itemize}
    \item The SIRST dataset encompasses 427 images, capturing a diverse array of environments including sky and maritime settings. It features a spectrum of infrared wavelengths, including both shortwave and mid-wave bands, as well as a specific 950 nm wavelength, catering to a broad spectrum of operational contexts.
    \item Comprising 1000 images, the IRSTD1K dataset offers a varied collection of targets such as drones, animals, boats, and vehicles, set against a backdrop of scenes ranging from urban to natural landscapes, including oceans, rivers, fields, mountains, and skies.
    \item The augmented SIRSTAUG dataset expands upon the original SIRST with 9070 images. It enhances target representation by cropping five images per target, positioned at different corners and the center, followed by random rotations at 0°, 45°, 90°, 135°, and 180° to simulate a variety of orientations.
    \item Lastly, the NUDT-SIRST dataset contains 1327 synthesized images, each derived from real-world scenarios peppered with diverse targets, against predominantly urban and natural scenes such as fields, oceans, and skies.
\end{itemize}

\subsubsection{\textbf{Implementation Details}} \label{subsubsec:implementation}

We employed Soft-IoU Loss \cite{SoftIoU} as the loss function, opted for AdamW \cite{AdamW} as the optimizer, and implemented a polynomial learning rate decay strategy in our approach. Additionally, we incorporated a deep supervision strategy for training the model.
The training was conducted on a system equipped with four RTX 3090 GPU using a distributed training approach. We trained four separate models on four different datasets. Due to significant variations in dataset distribution and resolution, we customized the hyperparameters for each dataset as shown in Tab.~\ref{tab:datasets}. 
The specific hyperparameter configurations of all compared methods are outlined in Tab.~\ref{tab:parameters}.

\begin{table*}[htbp]
  \setlength{\abovecaptionskip}{0cm}  %段前
  \renewcommand\arraystretch{1.2}
  \footnotesize
  \centering
  \vspace{-1\baselineskip}
\caption{Customed hyper-parameters for each dataset}
\label{tab:datasets}
% \vspace{-1\baselineskip}
\tabcolsep=0.3cm
\renewcommand\arraystretch{1.2}
\begin{center}
\begin{tabular}{ccccc | ccccc}
\toprule[1pt]
Dataset  & Epochs & Lr     & Batch & Resolution & Dataset  & Epochs & Lr     & Batch & Resolution\\ \hline
SIRST \cite{dai2021asymmetric}    & 1500   & 0.0001 & 4     & 512$\times$512 & IRSTD1K \cite{zhang2022isnet} & 500    & 0.0001 & 4     & 512$\times$512    \\
SIRSTAUG \cite{zhang2021agpcnet} & 500    & 0.0001 & 16    & 256$\times$256   & NUDT-SIRST  \cite{TIP2023DNANet}     & 1500   & 0.0001 & 16    & 256$\times$256 
   \\ \toprule[1pt]
\end{tabular}
\end{center}
  \vspace{-1\baselineskip}
\end{table*}

\subsubsection{\textbf{Evaluation Metrics}} \label{subsubsec:metric}
We used intersection over union (IoU), normalized intersection over union (nIoU) \cite{dai2021asymmetric}, probability of detection ($P_d$), and false alarm rate ($F_a$) as evaluation metrics for our experiment. They are defined as:
\begin{align}
    IoU &= \frac{TP}{T+P-TP}, \label{eq:IoU} \\
    nIoU &= \frac{1}{N}\sum_i^N\frac{TP(i)}{T(i)+P(i)-TP(i)}, \label{eq:nIoU} \\
    P_d &= \frac{TP}{TP+FN},\label{eq:Pd} \\
    F_a &= \frac{FP}{FP+TN}, \label{eq:Fa}
\end{align}
where $N$ is the total number of samples, $T$ and $P$ denote the number of ground truth and predicted positive pixels, respectively. $TP$, $FP$, and $FN$ denote the number of true positive, false positive, and false negative pixels, respectively.

\subsection{Ablation Study}

\subsubsection{\textbf{Ablation Study on Module-wise Performance Gain}}

In this study, we initiate our investigation from the fundamental U-Net architecture, which serves as our baseline. Our approach incrementally incorporates the novel components we propose: the DDC module, SeRank Module, and LSFF module. This systematic integration allows us to empirically validate the individual and collective enhancements these modules confer upon the detection performance.
It is imperative to underscore the significance of Positional Encoding (PE) within our framework. To this end, we have deconstructed the SeRank Module into two distinct entities: SeRank w/o PE and PE. This distinction is designed to facilitate the isolated assessment of the contributions made by the Top-K Selection mechanism and PE separately.

The experimental findings, as detailed in Table~\ref{tab:serankdet-ablation}, are revealing and substantiate several key insights:
1) The enhancement in detection performance attributed to each module is evident and operates independently, corroborating the motivation behind this study.
2) A comparative analysis between strategies (a) and (b) demonstrates that substituting vanilla convolution with our proposed DDC module yields a significant increase in the IoU metric, by 2.7\% on the SIRST dataset and 2.37\% on IRSTD1K.
3) Comparing strategies (a) and (c), we can clearly observe the performance improvement brought by the SeRank module. On the SIRST dataset, the introduction of the SeRank module boosts the IoU metric by 1.35\%.
4) Contrasting strategies (f) and (g) reveals that integrating the PE step into the SeRank w/o PE module, thus forming the complete SeRank module, significantly boosts the detection performance. This underscores the crucial role of preserving the positional information of feature points during the Top-K selection process.
5) Strategy (e) versus (g) comparison reveals that introducing a large receptive field selection mechanism into the cross-layer long-skip fusion framework substantially elevates detection capabilities.

\begin{table*}[htbp]
  \setlength{\abovecaptionskip}{0cm}  %段前
  \renewcommand\arraystretch{1.2}
  \footnotesize
  \centering
  \vspace{-1\baselineskip}
  \caption{Quantitative Ablation Study Assessing the Impacts of Core Components DDC, SeRank, LSFF, and PE within SeRankDet}
  \label{tab:serankdet-ablation}
  % \vspace{-2pt}
  \setlength{\tabcolsep}{8pt}
  \begin{tabular}{c|cccc|cc|cc|cc|cc}
      \multirow{2}{*}{Strategy} & \multicolumn{4}{c|}{Module} & \multicolumn{2}{c|}{SIRST}  & \multicolumn{2}{c|}{IRSTD1K}   & \multicolumn{2}{c|}{SIRSTAUG}   & \multicolumn{2}{c}{NUDT-SIRST} \\
      & DDC & SeRank w/o PE & LSFF & PE & IoU $\uparrow$ &  nIoU  $\uparrow$ & IoU $\uparrow$ &  nIoU $\uparrow$ & IoU $\uparrow$ &  nIoU $\uparrow$ & IoU $\uparrow$ &  nIoU $\uparrow$  \\
  \Xhline{1pt}
  % \vspace{.5\baselineskip}
(a) & \xmark & \xmark & \xmark & \xmark & 76.99  & 77.28 &  66.32 &  66.42 &  73.57 & 70.12  & 90.77  &  90.33 \\
  (b) & $\checkmark$ & \xmark & \xmark & \xmark & 79.46  & 77.82 &  68.69 &  68.43 &  75.26 & 71.45  & 92.65  &  91.88 \\
  (c) & \xmark & $\checkmark$ & \xmark  & $\checkmark$ & 78.34  & 78.12 & 69.94 & 68.08  & 75.03  & 70.73  & 93.20  &  92.55  \\
  (d) & \xmark & \xmark & $\checkmark$  & \xmark & 78.99  & 77.60 & 68.21 &  68.07 &  75.20 & 71.16  & 91.56  &  91.16  \\
  % \rowcolor[rgb]{0.9,0.9,0.9}
  (e) & $\checkmark$ & $\checkmark$ & \xmark  & $\checkmark$ & 80.64  & 78.75 & 70.90 &  68.71 &  75.74 & 71.50  & 92.78  &  92.73    \\
  (f) & $\checkmark$ & $\checkmark$ & $\checkmark$  & \xmark &  80.14 & 78.79 & 72.63 & 68.84  & 75.86  &  71.08 & 92.84  &  92.16  \\
  \hline
  \rowcolor[rgb]{0.9,0.9,0.9}
  (g) & \textbf{$\checkmark$} & \textbf{$\checkmark$} & $\checkmark$  & \textbf{$\checkmark$} & \textbf{81.27}  & \textbf{79.66} & \textbf{73.66} & \textbf{69.11}  &  \textbf{76.49} & \textbf{71.98}  &  \textbf{94.28} & \textbf{93.69} \\
  
\end{tabular}
% \vspace{-1\baselineskip}
\end{table*}

\begin{table*}[htbp]
  \setlength{\abovecaptionskip}{0cm}  %段前
  \renewcommand\arraystretch{1.2}
  \footnotesize
  \centering
  \vspace{-1\baselineskip}
  \caption{Ablation Study Results Highlighting the Impact of Different Convolution Types Constituting the DDC Module}
  \label{tab:ddc-ablation}
  % \vspace{-2pt}
  \setlength{\tabcolsep}{10pt}
  \begin{tabular}{c|ccc|cc|cc|cc|cc}
      \multirow{2}{*}{Strategy} & \multicolumn{3}{c|}{Module} & \multicolumn{2}{c|}{SIRST}  & \multicolumn{2}{c|}{IRSTD1K}   & \multicolumn{2}{c|}{SIRSTAUG}   & \multicolumn{2}{c}{NUDT-SIRST} \\
      & Conv & CDC & D.C. & IoU $\uparrow$ &  nIoU  $\uparrow$ & IoU $\uparrow$ &  nIoU $\uparrow$ & IoU $\uparrow$ &  nIoU $\uparrow$ & IoU $\uparrow$ &  nIoU $\uparrow$  \\
  \Xhline{1pt}
  % \vspace{.5\baselineskip}
  (a) & $\checkmark$ & \xmark & \xmark & 78.45  & 77.52 &  67.15 &  67.01 &  74.19 & 70.03  & 91.55  &  90.66 \\
  (b) & $\checkmark$ & $\checkmark$ & \xmark  & 78.81  & 77.57 & 67.75 & 67.85  & 75.04  & 70.65  & 92.12  &  91.82  \\
  % \rowcolor[rgb]{0.9,0.9,0.9}
  \hline
  \rowcolor[rgb]{0.9,0.9,0.9}
  (c) & $\checkmark$ & $\checkmark$ & $\checkmark$ & \textbf{79.46}  & \textbf{77.82} & \textbf{68.69} & \textbf{68.43}  &  \textbf{75.26} & \textbf{71.45}  &  \textbf{92.65} & \textbf{91.88} \\  
\end{tabular}
\end{table*}

\begin{table*}[htbp]
  \setlength{\abovecaptionskip}{0cm}  %段前
  \renewcommand\arraystretch{1.2}
  \footnotesize
  \centering
  \vspace{-1\baselineskip}
  \caption{Ablation Study on The Impact of Feature Point Cardinality in Top-K Selection on Detection Performance}
  \label{tab:optimal-k}
  % \vspace{-2pt}
  \setlength{\tabcolsep}{2pt}
  \begin{tabular}{c|c|c|cc|cc|cc|cc}
      \multirow{2}{*}{Strategy} & 
      \multirow{2}{*}{Top-K Operation} & 
      \multirow{2}{*}{Offset $o$} & 
      \multicolumn{2}{c|}{SIRST}  & \multicolumn{2}{c|}{IRSTD1K}   & \multicolumn{2}{c|}{SIRSTAUG}   & \multicolumn{2}{c}{NUDT-SIRST} \\
        &  &   & IoU $\uparrow$ &  nIoU  $\uparrow$ & IoU $\uparrow$ &  nIoU $\uparrow$ & IoU $\uparrow$ &  nIoU $\uparrow$ & IoU $\uparrow$ &  nIoU $\uparrow$  \\
  \Xhline{1pt}
  % \vspace{.5\baselineskip}
  (a) & $\operatorname{SparseAtt}(\mathbf{Q}, \mathbf{K}, \mathbf{V})=\operatorname{softmax}\left(\mathcal{T}_k\left(\frac{\mathbf{Q K}^{\top}}{\lambda}\right)\right) \mathbf{V}$ & /  & 80.26  & 78.77 &  72.20 &  68.65 & 75.30  &  70.73 & 93.01  & 92.65  \\
   % \cline{2-11}
   \hline
  (b) & \multirow{5}{*}{$\operatorname{SeRank}(\mathbf{X}) =  \left( 1 +  \operatorname{softmax}\left(\frac{\mathcal{T}_{\mathrm{p}}(\mathbf{X}) \bm{W}_Q \bm{W}_K^{\top} \mathcal{T}_{\mathrm{p}}(\mathbf{X})^{\top} - \mu}{\sigma + \varepsilon}\right) \right) \mathbf{X}$}
  & 0  & 80.86 & 79.07 &  72.41 &  68.79 &  75.32 & 71.25  & 92.67  &  92.24 \\
  (c) & & 1  & 80.81  & 79.10 & 72.78 & 68.56  & 75.88  & 71.75  & 93.83  &  93.04  \\
  (d) & & 2  & 80.16  & 78.39 & 72.55 & 68.03  & 76.25  & 71.79  & 94.02  &  93.13  \\
  % \cline{2-11}
  (e) & & 4  & 79.63  & 78.81 & 73.16 & 68.42  & 75.13  & 70.41  & 93.36  &  92.47  \\
  \rowcolor[rgb]{0.9,0.9,0.9} \textbf{(f)} & & 
   \textbf{3} & \textbf{81.27}  & \textbf{79.66} & \textbf{73.66} & \textbf{69.11} &  \textbf{76.49} & \textbf{71.98}  &  \textbf{94.28} & \textbf{93.69} \\
\end{tabular}
\end{table*}

\begin{figure*}[htbp]
  \centering
    \vspace{-1\baselineskip}
  \subfloat[SIRST]{
    \includegraphics[width=.4\textwidth]{
      "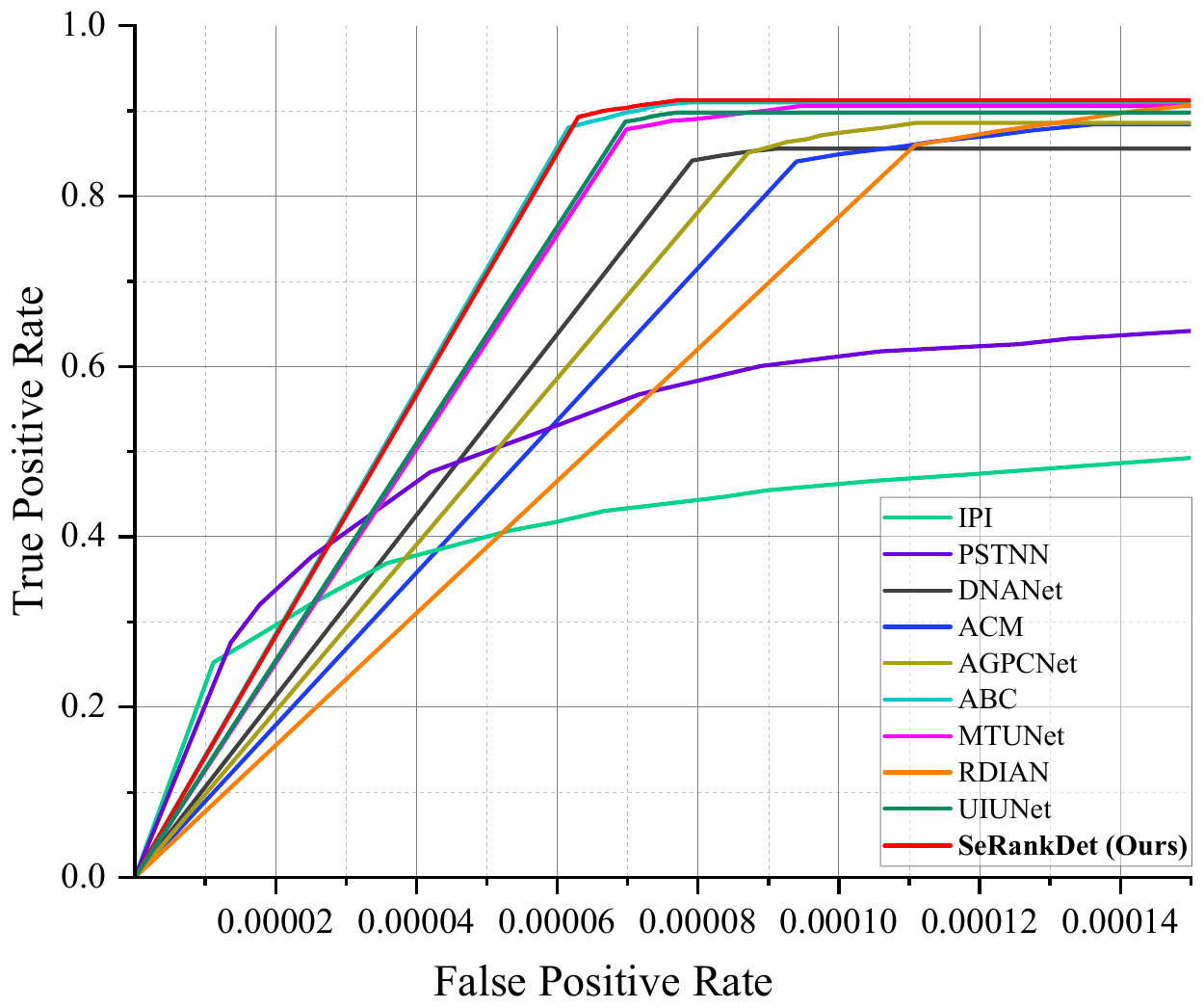"}
      \label{subfig:nuaa}
  }
  \subfloat[IRSTD1K]{
    \includegraphics[width=.4\textwidth]{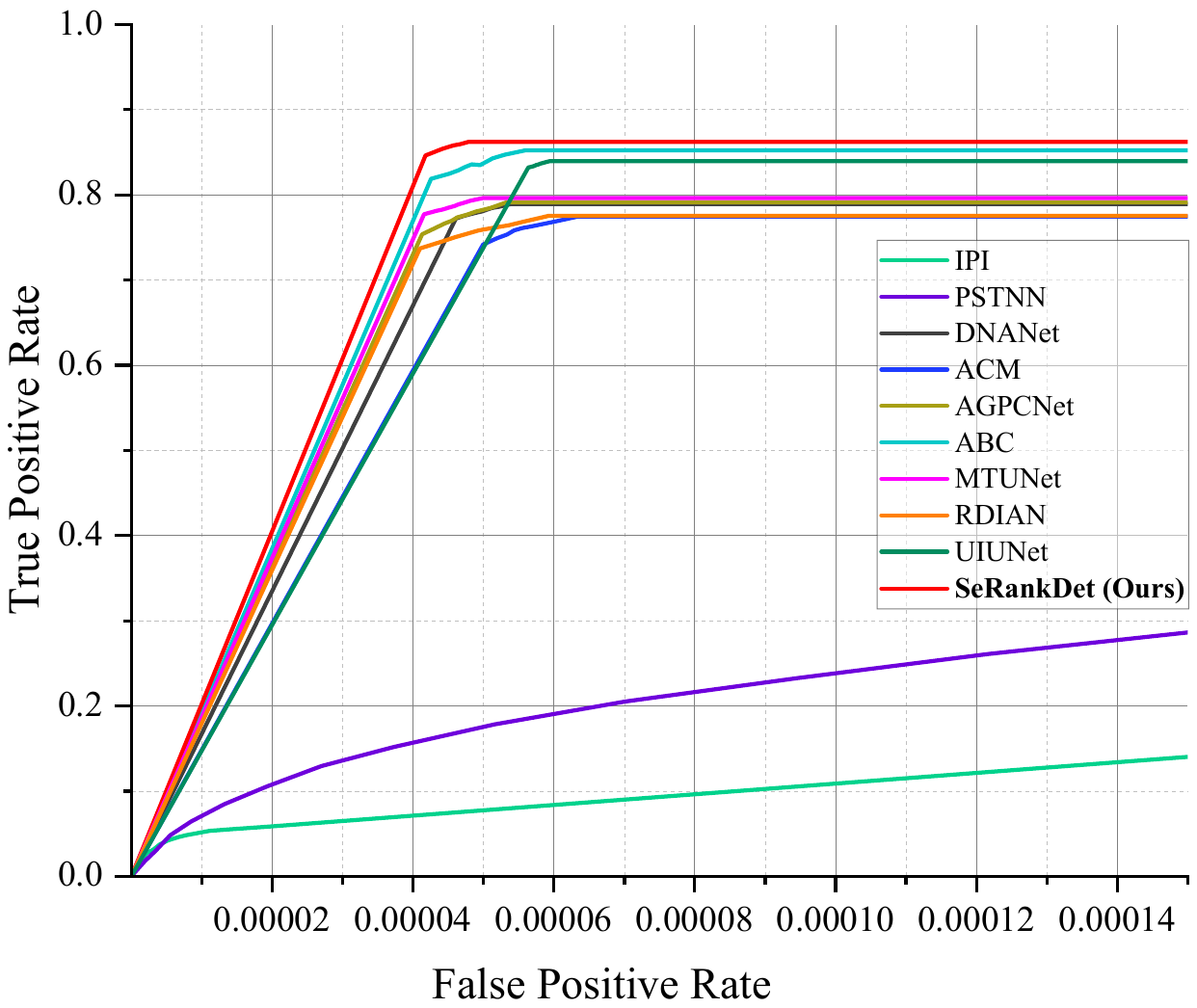}
      \label{subfig:IRSTDk}
  }
    % \vspace{-.5\baselineskip}
  \caption{Comparative ROC curve analysis of detection methods on (a) SIRST dataset and (b) IRSTD1K dataset. Our SeRankDet demonstrates superior performance, as evidenced by its consistently higher AUC values across both datasets, highlighting its robustness and adaptability in diverse scenarios.
  }
  \label{fig:roc}
  % \vspace{-1\baselineskip}
\end{figure*}

\subsubsection{\textbf{Ablation Study on Optimal Top-K Strategy}}

Our study primarily focuses on optimizing the application of the Top-K operation and determining the ideal number of feature points $K$ to be retained in this process. We achieve this by adjusting the offset $o$ in $K=2^{\lfloor\mathrm{log}_2(C)+o-2(i-1)\rfloor}$, wherein the detection performance, as illustrated in Table~\ref{tab:optimal-k}, initially improves and then decreases as $o$ increases.

This pattern implies that a lower count of selected feature points fails to incorporate sufficient background context, which is essential for the network to differentiate true targets from false alarms. On the contrary, an excessive number of selected feature points might dilute the target features during pooling, reducing their prominence. Therefore, we found that setting the offset to 3 yields the best results in our experiments.

Contrasting SeRank and SparseAttn, we observed a significant performance improvement with the former, reinforcing the effectiveness of our modifications. We attribute this enhancement to the application of Top-K prior to attention, which filters irrelevant background features, enabling the attention mechanism to focus more on the target region and thereby bolstering the discriminative power of feature representation. This modification also reduces the complexity of attention computation to a certain extent, thereby improving model efficiency.

\subsubsection{\textbf{Ablation Study on Synergy of Convolutions in DDC}}

Our DDC module is an advancement on vanilla convolution that amalgamates dilated convolution and central difference convolution (CDC) to yield a more robust feature representation.
It is imperative to dissect the individual contributions of each convolution type within this composite module.
Tab.~\ref{tab:ddc-ablation} reveals that, while the inclusion of CDC offers a performance increment for real-world datasets such as SIRST and IRSTD1K, the integration of dilated convolution significantly outpaces it. 
This suggests that although CDC unearths a higher number of potential targets, it concurrently introduces an abundance of false alarms, as exemplified in Fig.~\ref{fig:gallery}, limiting the actual performance gain.

In contrast, dilated convolution, with its expanded receptive field, demonstrates a potent ability to suppress these false alarms, substantially elevating the overall detection performance. This finding highlights the paramount importance of a large receptive field in the detection of infrared small targets, catalyzing our subsequent development of the LSFF module, which promises to further refine the detection process by leveraging extensive spatial context.

\subsection{Comparison with State-of-the-Arts} \label{subsec:sota}

\begin{table*}[h]
  \renewcommand\arraystretch{1.2}
  \footnotesize
  \centering
    % \vspace{-1\baselineskip}
  \caption{Comparison with Other State-of-the-art methods on four datasets.}
  \label{tab:sota}
  % \vspace{-1\baselineskip}
  \setlength{\tabcolsep}{3.5pt}
  \begin{tabular}{l|cccc|cccc|cccc|cccc}
    \multirow{2}{*}{Method}  & \multicolumn{4}{c|}{SIRST}  & \multicolumn{4}{c|}{IRSTD1K}   & \multicolumn{4}{c|}{SIRSTAUG}   & \multicolumn{4}{c}{NUDT-SIRST} \\
      &  IoU $\uparrow$ &  nIoU  $\uparrow$ & $P_d$ $\uparrow$ & $F_a$ $\downarrow$  & IoU $\uparrow$ &  nIoU $\uparrow$ & $P_d$ $\uparrow$ & $F_a$ $\downarrow$ & IoU $\uparrow$ &  nIoU $\uparrow$ & $P_d$ $\uparrow$ & $F_a$ $\downarrow$ & IoU $\uparrow$ &  nIoU $\uparrow$ & $P_d$ $\uparrow$ & $F_a$ $\downarrow$ \\
  \Xhline{1pt}
  \multicolumn{13}{l}{\textit{Background Suppression Methods}}  \\
  \hline
  % Bilateral TDLMS \cite{ZHAO201417} & 0.30 & 0.31 & 2.65 & 0.34 &
  NWMTH \cite{bai2010analysis} & 15.77 & 17.28 & 71.12 & 55.61 & 18.94 & 16.91 & 49.26 & 21.72 & 18.71 & 16.74 & 62,71 & 32.46 & 11.72 & 10.26 & 52.71 & 46.81 \\
  ILCM \cite{GRSL2014ILCM} & 18.27 & 19.70 & 38.58 & 40.26 & 13.05 & 11.46 & 43.55 & 47.32 & 5.25 & 6.44 & 26.27& 88.43 & 12.84 & 13.69 & 27.62 & 43.21 \\
  MPCM \cite{PR16MPCM} & 25.75 & 27.58 & 68.12 & 45.61 & 20.04 & 23.63 & 58.46 & 31.15 & 27.39 & 38.56 & 55.43 & 11.47 & 9.52 & 9.96 & 43.80 & 79.22 \\
  WLDM \cite{deng2016small} & 5.14 & 3.54 & 64.08 & 21.71 & 8.35 & 10.34 & 54.14 & 14.31 & 8.63 & 11.07 & 16.58 & 64.32 & 11.04 & 21.03 & 12.69 & 78.26 \\
  RLCM \cite{GRSL2018RLCM} & 25.45 & 27.78 & 86.68 & 62.31 & 16.92 & 21.03 & 70.36 & 67.60 & 8.40 & 11.90 & 70.53 & 31.44 & 18.37 & 18.11 & 79.57 & 68.96 \\
  FKRW \cite{qin2019infrared} & 22.06 & 28.08 & 81.77 & 16.32 & 10.39 & 16.25 & 69.54 & 24.37 & 10.05 & 16.56 & 72.42 & 66.87 & 12.67 & 21.73 & 79.51 & 67.13\\
  TLLCM \cite{han2019local} & 18.66 & 27.71 & 82.24 & 16.95 & 10.26 & 18.24 & 68.55 & 24.48 & 13.96 & 16.80 & 77.47 & 69.71 & 11.08 & 23.82 & 77.97 & 67.26\\
  GSWLCM \cite{qiu2022global} & 15.42 & 13.24 & 72.68 & 21.73 & 12.82 & 13.72 & 70.24 & 13.92 & 21.62 & 19.85 & 69.41 & 42.71 & 10.82 & 18.71 & 67.32 & 53.31 \\
  \hline
  \multicolumn{13}{l}{\textit{Low-rank and Sparse Decomposition}}  \\
  \hline
  IPI \cite{gao2013infrared} & 25.67&33.57&85.55&11.47&27.92&30.12&81.37&16.18&25.16&34.64&76.26&43.41&28.63&38.18&74.49&41.23\\
  NIPPS \cite{dai2017non} & 40.44 & 54.45 & 89.71 & 19.58 & 5.424  & 8.520 & 65.47  & 31.59 & 17.03  & 27.54 & 76.88 & 52.01  & 30.21  & 40.94  & 89.41  & 35.18\\
  RIPT \cite{dai2017reweighted}  & 11.05&19.91&79.08&22.61&14.11&17.43&77.55&28.31&24.13&33.98&78.54&56.24&29.17&36.12&91.85&344.3\\
  NRAM \cite{zhang2018infrared} & 25.25 & 31.25 & 75.47 & 21.47 & 9.882  & 18.71 & 72.48 & 24.73 & 8.972  & 15.27 & 71.47 & 68.98  & 12.08  & 18.61  & 72.58  & 84.77\\
  NOLC \cite{zhang2019infrared1} & 26.64 & 35.68 & 81.62 & 17.44 & 12.39  & 22.18 & 75.38 & 21.94 & 12.67  & 20.87 & 74.66 & 67.31  & 23.87  & 34.90  & 85.47  & 58.20\\
  PSTNN \cite{zhang2019infrared} & 22.40 & 29.59 & 77.95 & 29.11 & 24.57  & 28.71  & 71.99  & 35.26 & 19.14  & 27.16 & 73.14 & 61.58  & 27.72  & 39.80  & 66.13  & 44.17\\
  \hline
  \multicolumn{13}{l}{\textit{Deep Learning Methods}}  \\
  \hline
  ACM \cite{dai2021asymmetric} & 72.88  & 72.17 & 97.22 & 35.14 & 63.39  & 60.81  & 91.25  & 8.961 & 73.84  & 69.83 & 97.52 & 76.35 & 68.48  & 69.26  & 96.26  & 10.27\\
  AGPCNet \cite{zhang2021agpcnet} & 77.13  & 75.19 & 98.15 & 34.47 & 68.81  & 66.18  & 94.26  & 15.85 & 74.71  & 71.49 & 97.67 & 34.84 & 88.71  & 87.48  & 97.57  & 7.541\\
  DNANet \cite{TIP2023DNANet} &  75.55  & 75.90 & 99.08 & 13.48 & 68.87  & 67.53  & \textbf{94.95}  & 13.38 & 74.88  & 70.23 & 97.80 & 30.07 &  92.67  & 92.09  & \underline{99.5}3  & \underline{2.347} \\
  ISNet\cite{zhang2022isnet} & 80.02  & 78.12 & 99.18 & 4.924 &  68.77  & 64.84  & 95.56  & 15.39 &72.50 & 70.84 & 98.41 & 28.61 & 89.81 & 88.92 & 99.12 & 4.211\\
  UIUNet \cite{wu2022uiu} & 80.08  & 78.09 & \underline{100.0} & 4.040 &  69.13  & 67.19  & 94.27& 16.47 &74.24&70.57&98.35&23.13&90.77&90.17&99.29&2.390\\
  RDIAN \cite{sun2023receptive} & 72.85  & 73.96 & 99.08 & 28.48 & 64.37  & 64.90  & 92.26  & 18.20 & 74.19  & 69.80 & \underline{99.17} &23.97 & 81.06  & 81.72  & 98.36  & 14.09\\
  MTUNet \cite{wu2023mtu} & 78.75  & 76.82 & 98.17 & 8.650 &  67.50  & 66.15  & 93.27  & 14.75 &74.70&70.66&98.49&39.73&87.49&87.70&98.60&3.760\\
  ABC \cite{pan2023abc} & \underline{81.01} & \underline{79.00} & \underline{100.0} & \underline{2.440} & \underline{72.02} & \underline{68.81} & 93.76& \underline{9.457} &76.12&\underline{71.83}& \textbf{99.59} & \underline{20.33} &\underline{92.85}&\underline{92.45}& 99.29 & 2.900 \\
  \hline
  \rowcolor[rgb]{0.9,0.9,0.9}$\star$ \textbf{SeRankDet (Ours)}  & \textbf{81.27} & \textbf{79.66} & \textbf{100.0} & \textbf{1.860} & \textbf{73.66} & \textbf{69.11} & \underline{94.28} & \textbf{8.370} & \textbf{76.49} & \textbf{71.98} & \underline{99.17} & \textbf{18.90} & \textbf{94.28} & \textbf{93.69} & \textbf{99.77} & \textbf{2.030}\\
\end{tabular}
\end{table*}

\begin{table}[h]
  \renewcommand\arraystretch{1.2}
  \footnotesize
  \centering
  \caption{Performance comparison of different methods on the NoisySIRST dataset with varying levels of Gaussian white noise.}
  \label{tab:noise}
  \setlength{\tabcolsep}{3.5pt}
  \begin{tabular}{l|cc|cc|cc}
    \multirow{2}{*}{Method}  & \multicolumn{2}{c|}{$\sigma_n=10$}  & \multicolumn{2}{c|}{$\sigma_n=20$}   & \multicolumn{2}{c}{$\sigma_n=30$} \\
      &  IoU $\uparrow$ &  nIoU  $\uparrow$ & IoU $\uparrow$ &  nIoU $\uparrow$ & IoU $\uparrow$ &  nIoU $\uparrow$ \\
  \Xhline{1pt}

  ACM \cite{dai2021asymmetric} & 73.71 & 71.70 & 69.66 & 68.24 & 66.95 & 65.38 \\
  AGPCNet \cite{zhang2021agpcnet} & 71.41 & 70.96 & 71.30 & 69.90 & 68.05 & 67.28 \\
  DNANet \cite{TIP2023DNANet} & 71.44 & 72.57 & 70.53 & 68.14 & 68.50  & 66.42 \\
  UIUNet \cite{wu2022uiu} & 77.01 & 74.45 & 73.36 & 70.44 & 69.16 & 69.17 \\
  RDIAN \cite{sun2023receptive}  & 70.86 &  71.57 & 70.82 & 68.40 & 66.09 & 66.01 \\  
   MTUNet \cite{wu2023mtu} & 75.42 & 73.11 & 71.27 & 69.59 & 66.40 & 66.94 \\
  ABC \cite{pan2023abc}  & 77.33  & 74.97  &  73.01 & 70.15  &  69.24 & 68.93 \\
  \hline
  \rowcolor[rgb]{0.9,0.9,0.9}$\star$ \textbf{SeRankDet (Ours)} & \textbf{77.59} & \textbf{75.30} & \textbf{73.49} & \textbf{71.94} & \textbf{69.68} & \textbf{69.68} \\
\end{tabular}
\end{table}

Our SeRankDet method was benchmarked against current leading or state-of-the-art methods on four datasets, using metrics such as IoU, nIoU, $P_d$, and $F_a$. The experimental outcomes are presented in Tab.~\ref{tab:sota}.
It can be see that:
1) Our SeRankDet achieved the best performance across all metrics, underscoring the efficacy of our ``Pick of the Bunch'' strategy. This approach entails using a high-sensitivity feature extractor to preserve dim targets and a robust module to eliminate false alarms.
2) A comparison reveals that deep learning methods significantly outperform traditional ones. This disparity highlights the inadequacy of model-driven methods, reliant on global sparsity and contrast priors, for complex image scenarios, while data-driven approaches maintain high detection accuracy. The success of data-driven methods is attributed to their ability to extract semantic information and their minimal hyperparameter dependence, enhancing robustness against scene variations.
3) Our method consistently outperformed the ABC approach. ABC's effective integration of global and local information accounts for its commendable performance. However, SeRankDet, with its precision in target localization, advanced feature extraction, and dynamic fusion modules, ensures finer segmentation of infrared small targets, leading to superior outcomes.

The ROC curves for the methods on SIRST and IRSTD1K datasets are depicted in Fig.~\ref{fig:roc}. The ROC analysis corroborates the findings in Tab.~\ref{tab:sota}, confirming the stability and superiority of deep learning methods, particularly SeRankDet, over traditional techniques.

To further assess the robustness of our method under noisy conditions, we constructed the NoisySIRST dataset by adding varying levels of Gaussian white noise ($\sigma = 10, 20, 30$) to the original SIRST dataset. This simulates real-world scenarios where infrared images are often degraded by severe noise interference. For reproducibility, we have made the NoisySIRST dataset publicly available\footnote{\url{https://github.com/GrokCV/SeRankDet}}.

\begin{figure*}[htbp]
    \centering
    \includegraphics[width=.98\linewidth]{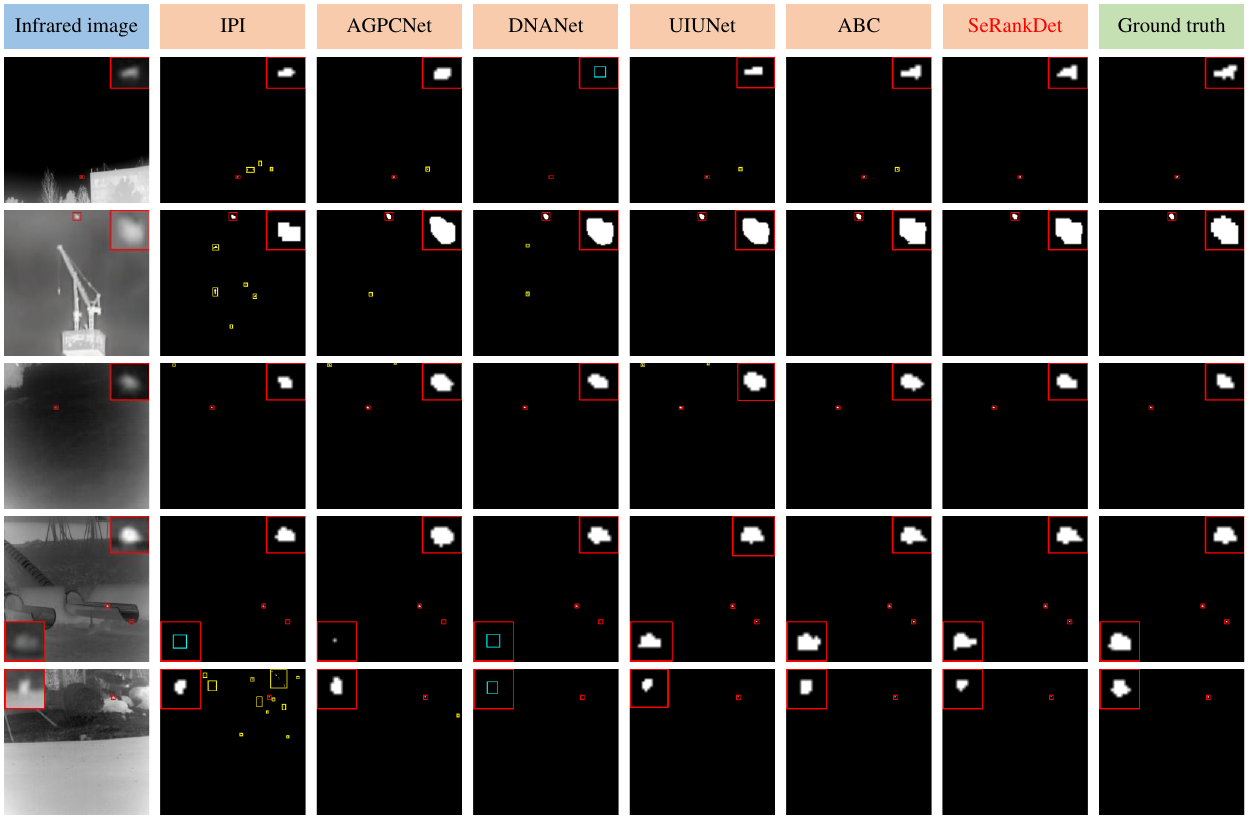}
    \caption{
    Visualization comparison of detection results via different methods on representative images from SIRST and IRSTD1K datasets. The red, yellow, and cyan boxes denote correct detections, false alarms, and missed detections, respectively.}
    \label{fig:visual}
\end{figure*}
\begin{figure*}[htbp]
    \centering
    \includegraphics[width=.98\linewidth]{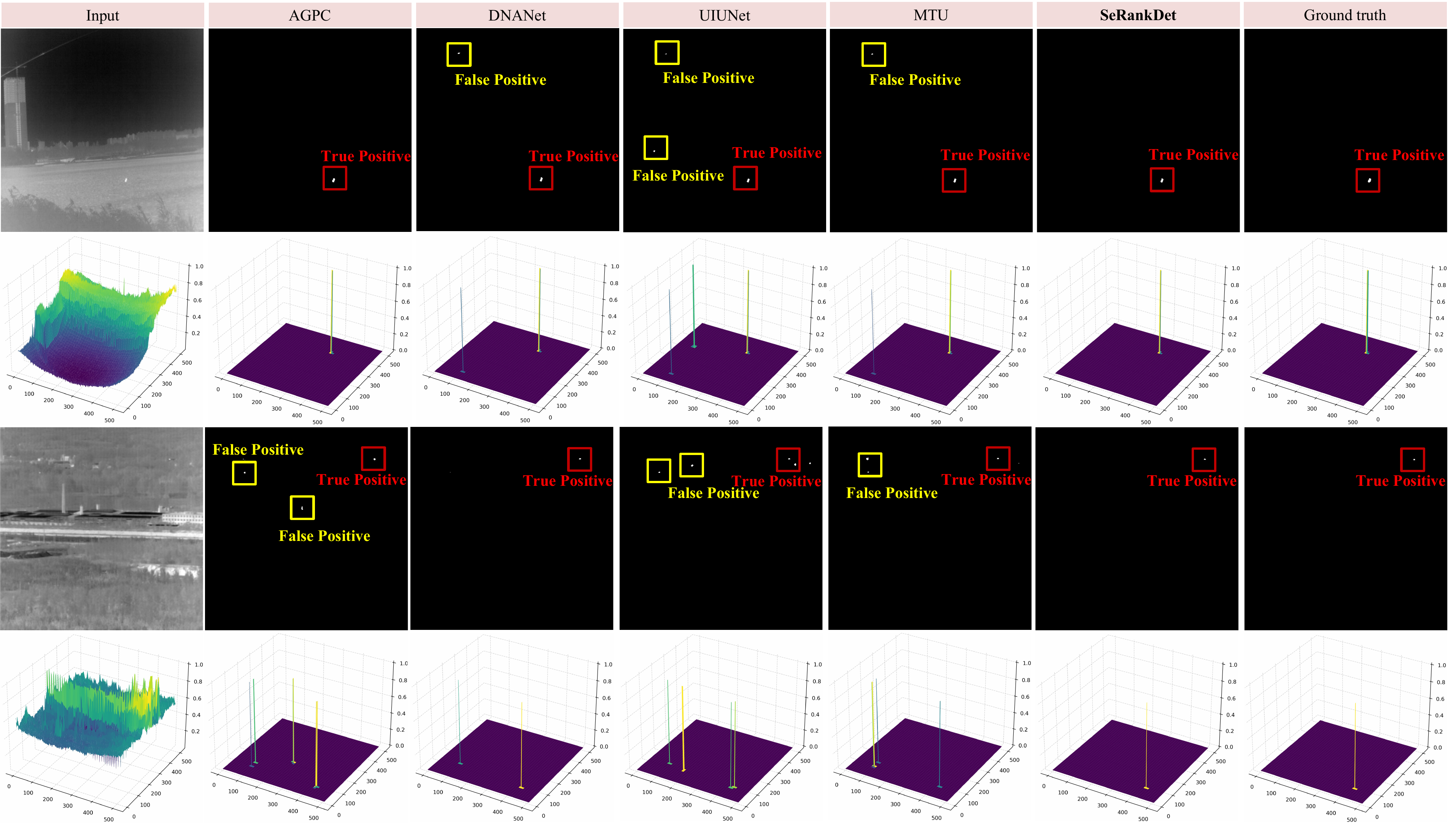}
    \caption{
    3D visualization comparison of detection results via different methods. The red, yellow, and cyan boxes denote correct detections and false alarms.}
    \label{fig:3d}
\end{figure*}

\begin{figure*}[htbp]
    \centering
    \includegraphics[width=.98\linewidth]{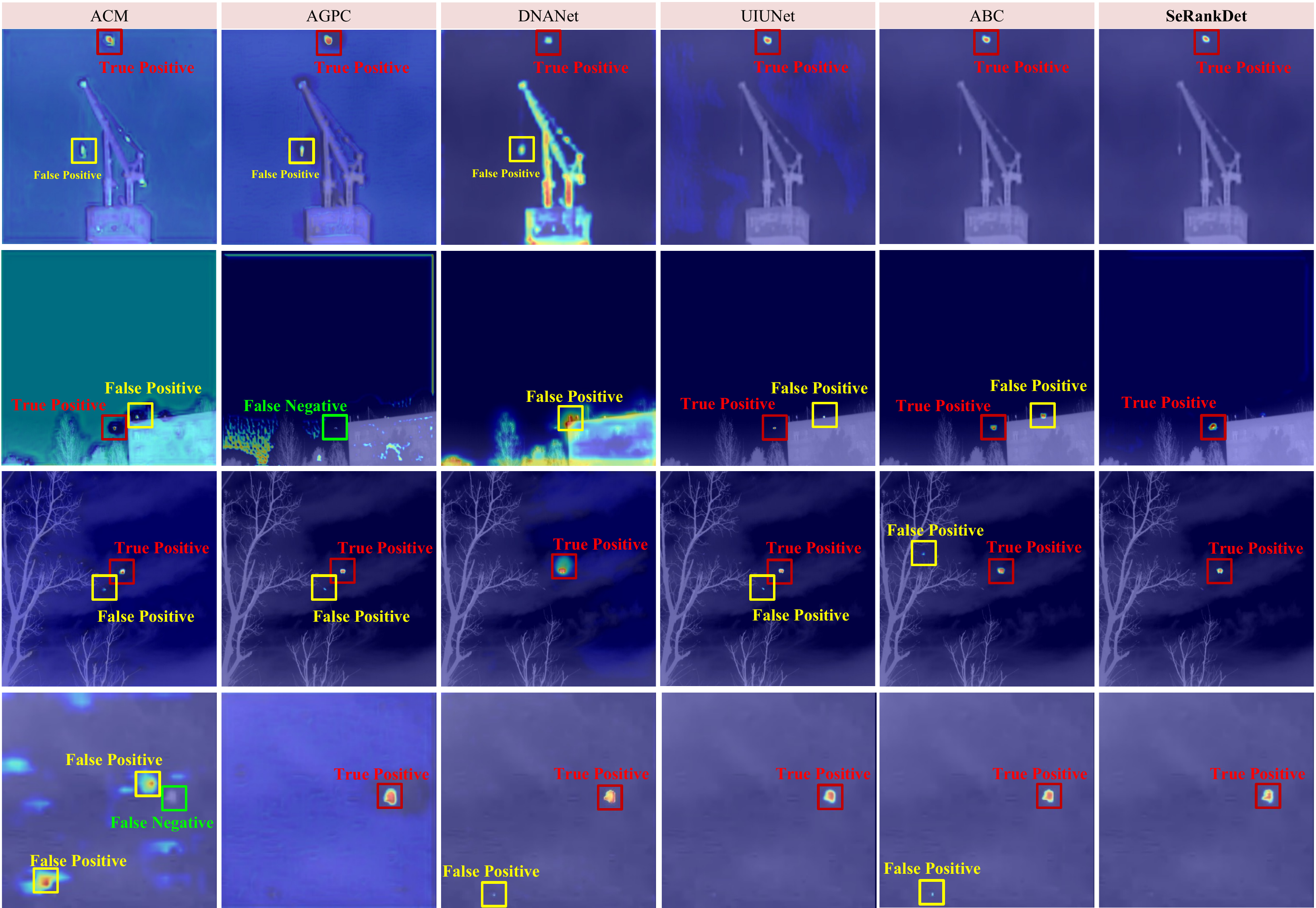}
    \caption{Feature Visualization Comparison. True positive detections, false alarms, and missed detections are denoted by red, yellow, and green bounding boxes, respectively. The proposed SeRankDet exhibits superior performance, as evidenced by its higher true positive detections and fewer false alarms and missed detections compared to other methods. }
    \label{fig:gradcam}
\end{figure*}

Upon performing experiments on this newly created NoisySIRST dataset, we compared the performance of our proposed SeRankDet method with other cutting-edge approaches. A comprehensive summary of these comparative results is provided in Table \ref{tab:noise}. The results clearly illustrate that our SeRankDet method outperforms other state-of-the-art approaches across all noise levels in terms of IoU and nIoU metrics, thereby highlighting its superior noise robustness. Notably, our method's performance advantage under substantial noise interference can be ascribed to our Top-K operation. This operation effectively extracts the key features of the target, thereby preventing the dilution of weak target features by background features in operations like Linear Squeeze. As a result, it ensures the detection accuracy of the target even in challenging noisy conditions.

\subsection{Visualization Analysis}

The qualitative results presented in Fig.~\ref{fig:visual}, Fig.~\ref{fig:3d}, and Fig.~\ref{fig:gradcam} offer a visual comparison of the detection performance of various methods on representative images from the SIRST and IRSTD1K datasets. Traditional methods generate significantly more false alarms (yellow boxes) and missed detections (cyan boxes) compared to deep learning-based approaches, highlighting the advantages of deep learning techniques in infrared small target detection. Deep neural networks can learn rich, hierarchical features from the data, enabling them to effectively distinguish between true targets and background clutter.
Among the deep learning-based methods, our proposed approach, which incorporates the novel SeRank Module and large selective feature fusion, consistently outperforms other state-of-the-art methods.

While UIUNet and ABC demonstrate competitive performance, their results still lag behind our method in certain cases. As shown in Fig.~\ref{fig:visual} and Fig.~\ref{fig:3d}, UIUNet occasionally generates false alarms in complex backgrounds, such as in the first row. In contrast, our method maintains high detection accuracy and robustness across diverse scenes, benefiting from the SeRank Module's ability to adapt to varying target and background characteristics.

The GradCAM visualizations in Fig.~\ref{fig:gradcam} further illustrate the superiority of our approach. Our method generates more focused and accurate activation maps compared to UIUNet and ABC, indicating its stronger ability to localize and highlight the true targets. This visual evidence aligns with the quantitative results, confirming the effectiveness of the proposed SeRank Module and large selective feature fusion mechanism in infrared small target detection.

% !TEX root = ../main.tex
% \bibliography{../reference.bib}

% !TEX root = ../main.tex

\section{Conclusion} \label{sec:conclusion}

In conclusion, this work addressed the challenging problem of infrared small target detection amidst intricate background clutter.
Our proposed solution, SeRankDet, transcends the hit-miss trade-off and sets new standards for detection accuracy, guided by the ``Pick of the Bunch'' principle.
The core of SeRankDet is the SeRank module, which implements a non-linear Top-K selection mechanism to preserve vital target signals with constant complexity.
Complementing the SeRank module is our large selective feature fusion strategy, which allows SeRankDet to adjust the integration of features in an adaptive manner.
Finally, the DDC module further refines the network's discernment by merging differential convolution with dilated convolution, which together markedly heighten the distinction between targets and their visually similar counterparts.
Despite its simple architecture, SeRankDet establishes new benchmarks on multiple public datasets, confirming its superior performance for infrared small target detection.

\bibliographystyle{IEEEtran}
\bibliography{./reference.bib}

% !TEX root = ../main.tex

% Yimian Dai
\begin{IEEEbiography}[{\includegraphics[width=1in,height=1.25in,clip,keepaspectratio]{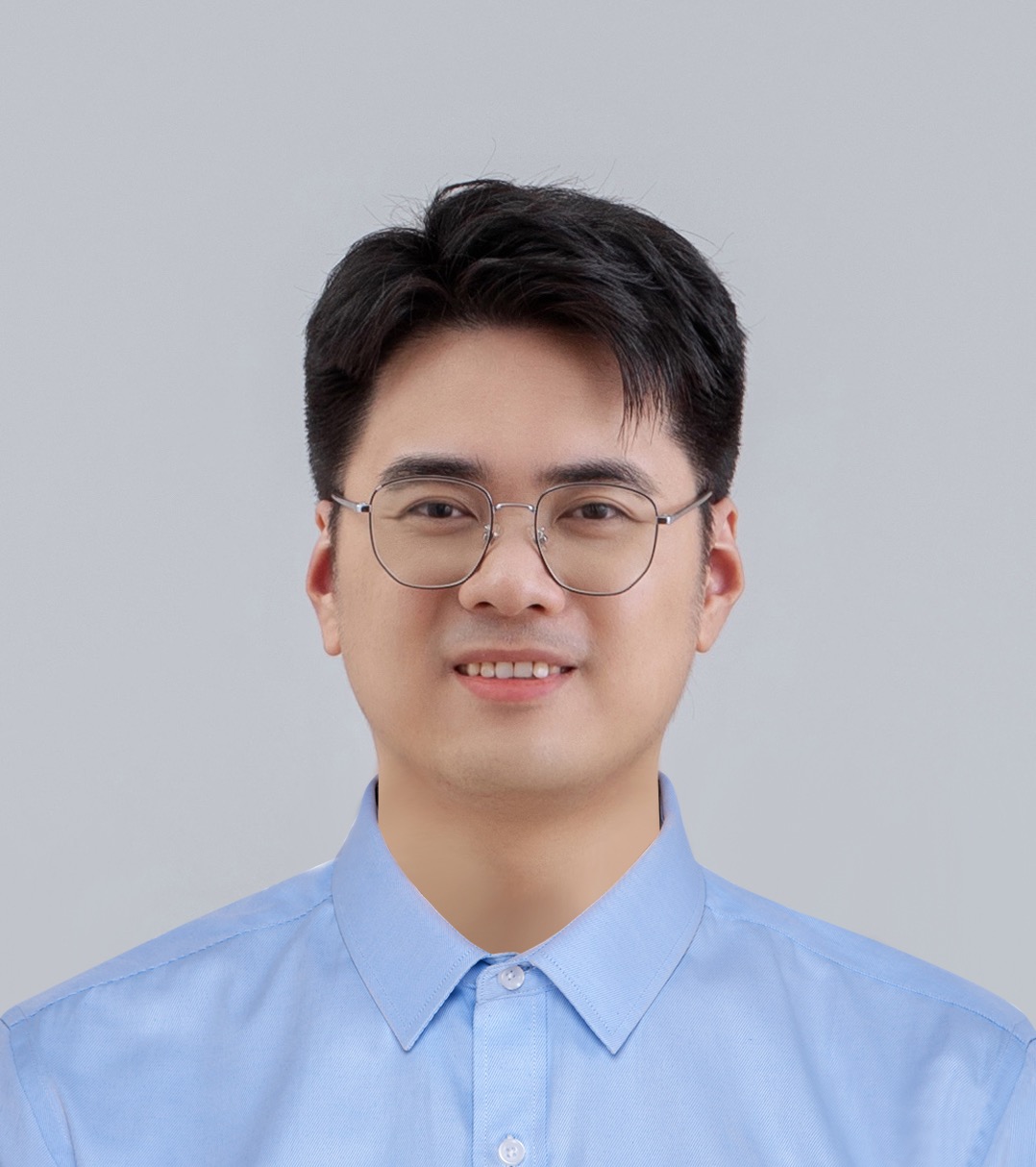}}]{Yimian Dai} received the Ph.D. degree in signal and information processing from Nanjing University of Aeronautics and Astronautics, Nanjing, China, in 2021, during which he honed his research skills as a visiting Ph.D. student at the University of Copenhagen and the University of Arizona between March 2018 and October 2020. Since 2021, he has been with the School of Computer Science and Engineering, Nanjing University of Science and Technology (NJUST), Nanjing, where he is currently an Assistant Researcher. His research interests include remote sensing, computer vision, and deep learning, with a focus on developing algorithms for object detection, data assimilation, and computational imaging to tackle real-world challenges. He has authored more than 20 peer-reviewed journal and conference papers such as IJCV, TGRS, TAES, etc.
\end{IEEEbiography}

% 培文
\begin{IEEEbiography}[{
\includegraphics[width=1.45in,height=1.3in,clip,keepaspectratio]{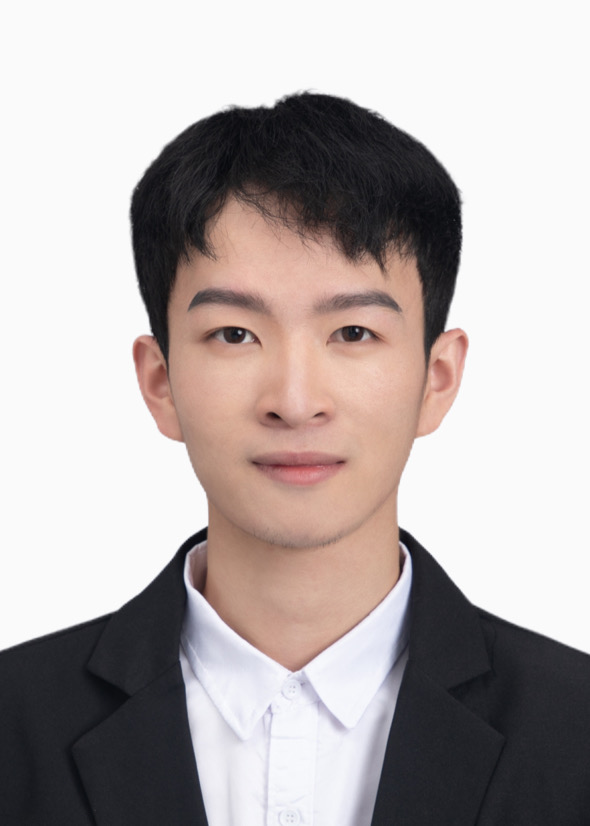}}]{Peiwen Pan} is currently Master student at the School of Computer Science and Engineering, Nanjing University of Science and Technology, Nanjing China. He received the B.S. degree in computer science and technology from the Changsha University of Science and Technology, Changsha, China. He won the first place in the 2022 Second National Information Fusion Challenge. His research direction is infrared small target detection and segmentation, and recommendation algorithms.
\end{IEEEbiography}

% Yulei Qian
\begin{IEEEbiography}[{\includegraphics[width=1.1in,height=1.35in,clip,keepaspectratio]{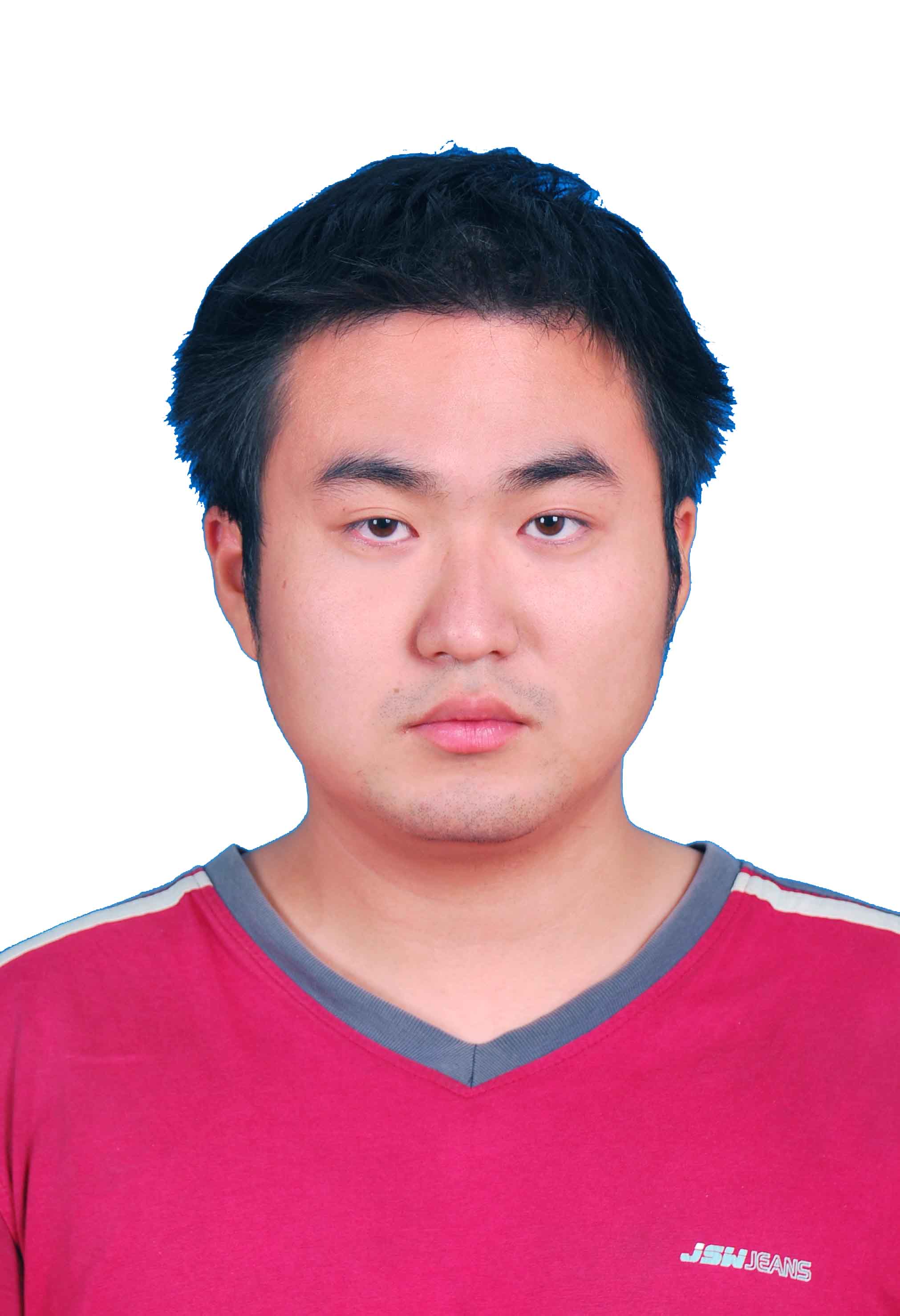}}]{Yulei Qian} received his PhD degree from College of Electronic and Information Engineering, Nanjing University of Aeronautics and Astronautics in 2020. He is now an engineer in Nanjing Marine Radar Institute. His main research interests include target detection, radar imaging, satellite synthetic aperture radar, sparse recovery, deconvolution and radar waveform design. He has published 10+ paper in remote sensing journals and conferences.
\end{IEEEbiography}

% Yuxuan Li
\begin{IEEEbiography}[{\includegraphics[width=1.1in,height=1.35in,clip,keepaspectratio]{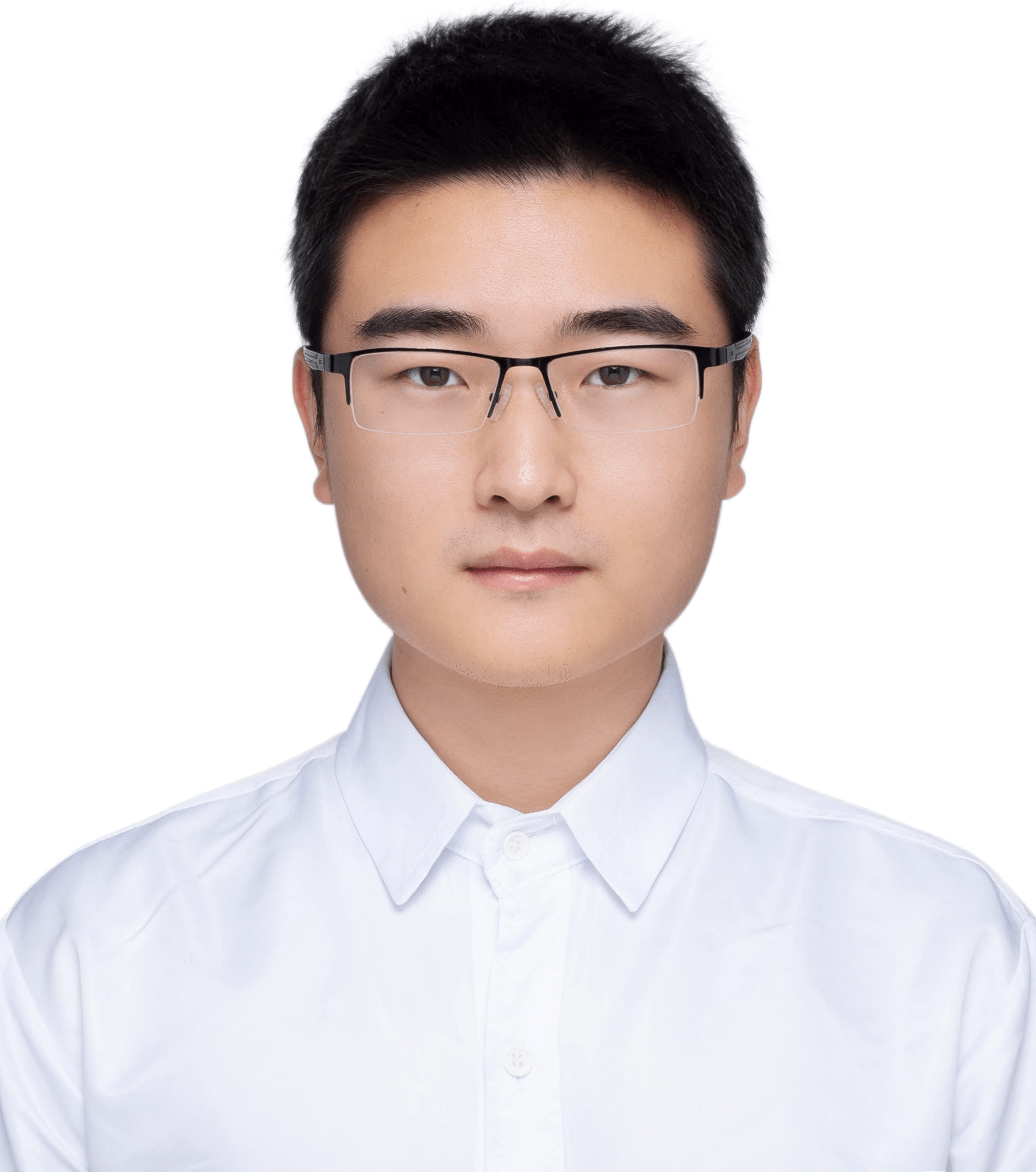}}]{Yuxuan Li} is currently a Ph.D. student at the Department of Computer Science, Nankai University, China.
  He graduated from University College London (UCL) with a first-class degree in Computer Science. He was the champion of the Second Jittor Artificial Intelligence Challenge in 2022, was awarded 2nd place in Facebook Hack-a-Project in 2019 and was awarded 2nd place in the Greater Bay Area International Algorithm Competition in 2022. His research interests include neural architecture design, and remote sensing object detection. 
\end{IEEEbiography}

% Xiang Li
\begin{IEEEbiography}[{\includegraphics[width=1.1in,height=1.35in,clip,keepaspectratio]{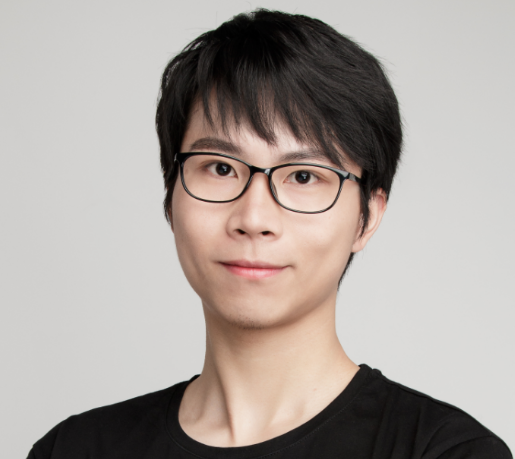}}]{Xiang Li} is an Associate Professor in College of Computer Science, Nankai University. He received the PhD degree from the Department of Computer Science and Technology, Nanjing University of Science and Technology (NJUST) in 2020. There, he started the postdoctoral career in NJUST as a candidate for the 2020 Postdoctoral Innovative Talent Program. In 2016, he spent 8 months as a research intern in Microsoft Research Asia, supervised by Prof. Tao Qin and Prof. Tie-Yan Liu. He was a visiting scholar at Momenta, mainly focusing on monocular perception algorithm. His recent works are mainly on: neural architecture design, CNN/Transformer, object detection/recognition, unsupervised learning, and knowledge distillation. He has published 20+ papers in top journals and conferences such as TPAMI, CVPR, NeurIPS, etc.
\end{IEEEbiography}

% Jian Yang
\begin{IEEEbiography}[{
  \includegraphics[width=1.45in,height=1.3in,clip,keepaspectratio]{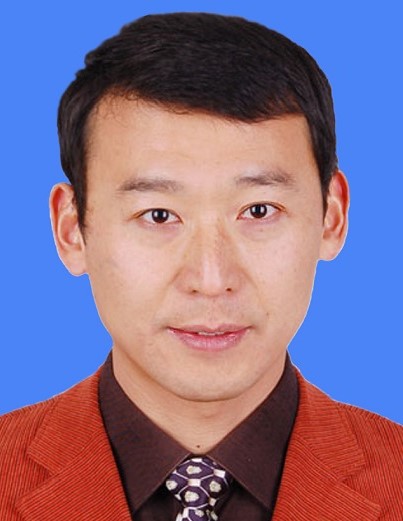}}]{Jian Yang} received the PhD degree from Nanjing University of Science and Technology (NJUST) in 2002, majoring in pattern recognition and intelligence systems. From 2003 to 2007, he was a Postdoctoral Fellow at the University of Zaragoza, Hong Kong Polytechnic University and New Jersey Institute of Technology, respectively. From 2007 to present, he is a professor in the School of Computer Science and Technology of NJUST. Currently, he is also a visiting distinguished professor in the College of Computer Science of Nankai University. He is the author of more than 300 scientific papers in pattern recognition and computer vision. His papers have been cited over 40000 times in the Scholar Google. His research interests include pattern recognition and computer vision. Currently, he is/was an associate editor of Pattern Recognition, Pattern Recognition Letters, IEEE Trans. Neural Networks and Learning Systems, and Neurocomputing. He is a Fellow of IAPR. 
\end{IEEEbiography}

% Huan Wang
\begin{IEEEbiography}[{
\includegraphics[width=1.45in,height=1.3in,clip,keepaspectratio]{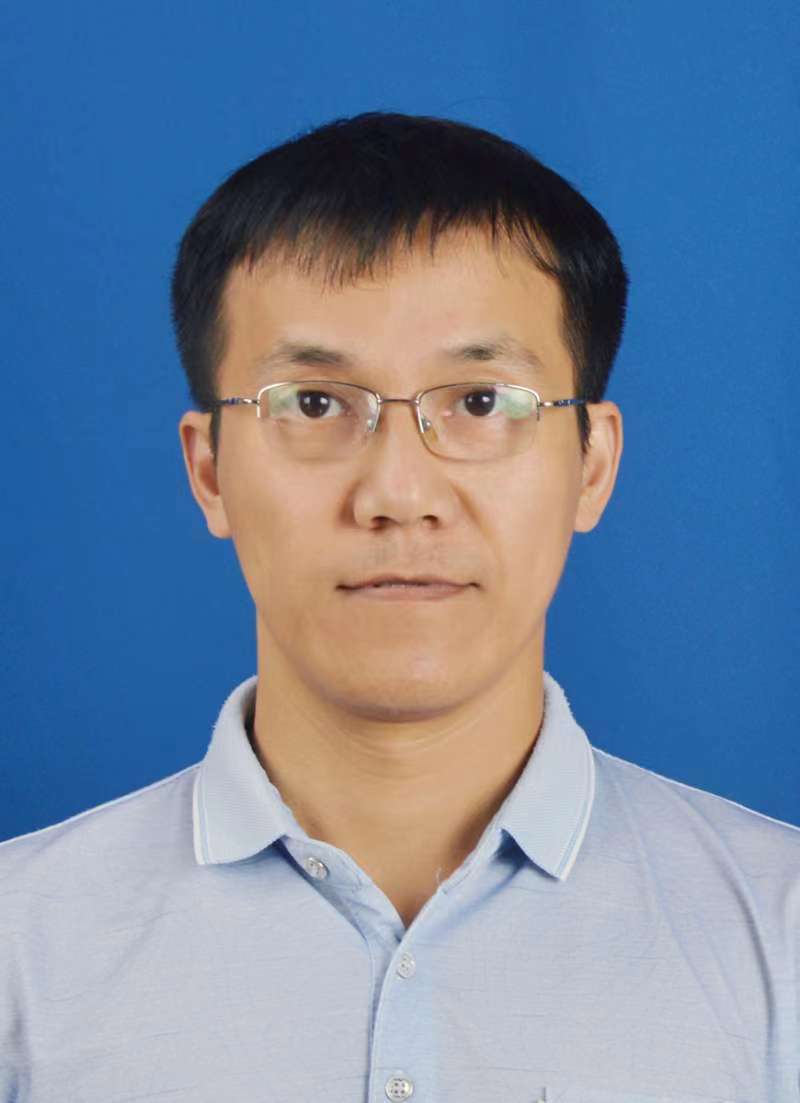}}]{Huan Wang} received B.S. degree in computer science from the Jiangsu university of Science and Technology, Zhen Jiang, China. He received the Doctor degree in pattern recognition and intelligent systems from Nanjing University of Science and Technology (NJUST), Nanjing, China. He is currently an associate professor with the School of Computer Science and Engineering, NJUST. He has authored over 40 scientific papers in image processing, computer vision, pattern recognition, and artificial intelligence. His current research interests include pattern recognition, robot vision and image processing.
\end{IEEEbiography} % 作者介绍

\end{document}